\def\year{2021}\relax
\documentclass[letterpaper]{article} 
\usepackage{aaai21}  
\usepackage{times}  
\usepackage{helvet} 
\usepackage{courier}  
\usepackage[hyphens]{url}  
\usepackage{graphicx} 
\usepackage{graphicx}  
\usepackage{natbib}  
\usepackage{caption} 
\usepackage{amsmath}
\usepackage{amssymb}
\usepackage{adjustbox}
\usepackage{makecell}
\usepackage[switch]{lineno}
\usepackage{subcaption}
\usepackage{threeparttable}
\nocopyright

\title{Towards Domain Invariant Single Image Dehazing}

\author {
        Pranjay Shyam\textsuperscript{\rm 1} \quad
        Kuk-Jin Yoon \textsuperscript{\rm 2} \quad 
        Kyung-soo Kim \textsuperscript{\rm 1} \\
}
\affiliations {
    \textsuperscript{\rm 1} Mechatronics, Systems and Control Lab \quad \textsuperscript{\rm 2} Visual Intelligence Lab \\
    Department of Mechanical Engineering\\
    Korea Advanced Institute of Science and Technology (KAIST)\\
    Daejeon, Republic of Korea, 34141\\
    \{pranjayshyam, kjyoon, kyungsookim\}@kaist.ac.kr
}

\begin{document}
\maketitle

\begin{abstract}
Presence of haze in images obscures underlying information, which is undesirable in applications requiring accurate environment information. To recover such an image, a dehazing algorithm should localize and recover affected regions while ensuring consistency between recovered and its neighboring regions. However owing to fixed receptive field of convolutional kernels and non uniform haze distribution, assuring consistency between regions is difficult. In this paper, we utilize an encoder-decoder based network architecture to perform the task of dehazing and integrate an spatially aware channel attention mechanism to enhance features of interest beyond the receptive field of traditional conventional kernels. To ensure performance consistency across diverse range of haze densities, we utilize greedy localized data augmentation mechanism. Synthetic datasets are typically used to ensure a large amount of paired training samples, however the methodology to generate such samples introduces a gap between them and real images while accounting for only uniform haze distribution and overlooking more realistic scenario of non-uniform haze distribution resulting in inferior dehazing performance when evaluated on real datasets. Despite this, the abundance of paired samples within synthetic datasets cannot be ignored. Thus to ensure performance consistency across diverse datasets, we train the proposed network within an adversarial prior-guided framework that relies on a generated image along with its low and high frequency components to determine if properties of dehazed images matches those of ground truth. We preform extensive experiments to validate the dehazing and domain invariance performance of proposed framework across diverse domains and report state-of-the-art (SoTA) results. The source code with pretrained models will be available at https://github.com/PS06/DIDH.
\end{abstract}

\section{Introduction}


Visibility degradations arising from environmental variations such as haze, smoke and fog affect image quality by concealing underlying information which is undesirable in applications where accurate surrounding information is necessary for safe operation such as autonomous vehicles, aerial robots and intelligent infrastructure. To overcome complications arising from deteriorations such as haze and fog, image dehazing has been extensively studied to recover clean image from its degraded version. Common approaches rely upon haze estimation using atmospheric scattering model \cite{mccartney1976optics, narasimhan2000chromatic, narasimhan2002vision}(Eq. \ref{eq:eq1}) to recover haze affected regions, that establishes a pixel-wise $(x)$ relationship between ambient light intensity $(A)$ and transmission matrix $t(x) = e^{-\beta d(x)}$ (representing fraction of light reaching camera sensor) by using scene depth $(d(x))$ and scattering coefficient $(\beta)$, to generate a hazy image $(I(x))$ from a clean image $(J(x))$. 

\begin{equation}
    I(x) = J(x)t(x) + A(1 - t(x))
    \label{eq:eq1}
\end{equation}

Traditional computer vision based dehazing algorithms relied upon handcrafted priors such as dark channel \cite{he2010single}, color attenuation \cite{zhu2015fast}, bi-channel \cite{jiang2017image} and color lines \cite{fattal2014dehazing, Berman_2016_CVPR} to estimate atmospheric light or transmission map to recover dehazed image, following atmospheric scattering model. However strong reliance on priors makes these methods vulnerable in scenarios when these priors don't hold. To avoid dependence on priors, different models leveraging the feature extraction capabilities of convolutional neural networks (CNNs) were proposed, following either atmospheric scattering model \cite{he2010single, cai2016dehazenet, zhang2018densely} or end-to-end approach \cite{ren2018gated, li2017all, mei2018progressive, chen2019gated, engin2018cycle} to estimate dehazed images. 

Although learning based approaches represent current state-of-the-art (SoTA), they require large number of training samples accurately representing haze scenarios in different outdoor settings. Constructing such a large scale real dataset is both expensive and time consuming, thus atmospheric scattering model is used to generate synthetic haze corresponding to a clean image. However such approaches are limited in considering the effect of airborne particles on different wavelengths \cite{li2020learning} apart from presence of wind, resulting in difference between real and synthetic images in form of domain difference and haze distribution. This results in performance gap arising between real and synthetic haze trained models when evaluated upon either of the datasets. 

To overcome the dual challenge of varying haze distribution and domain difference. In this paper, we propose a framework for achieving domain and distribution invariant dehazing framework. We begin by focusing on achieving consistent performance irrespective of haze distribution and highlight the importance of localizing haze affected regions as a necessary step towards effective dehazing. To attain such characteristics, we concentrate upon data augmentation and architecture of the underlying CNN. Specifically, to generate non-uniform haze distribution on synthetic samples, we leverage greedy localized data augmentation proposed in \cite{shyam2020copy} that copies multiple patches of varying shapes and sizes from noisy image onto corresponding paired clean image to generate non-uniform noise patches. For our purpose, this approach results in generation of non-homogeneous haze. In order to accurately recover image regions affected by unknown haze distribution, we utilize an encoder-decoder framework built upon UNet \cite{ronneberger2015u}, to aggregate and represent features across different scales into a higher order latent space which is subsequently be used by a decoder to reconstruct haze free image. To ensure information and color consistency between recovered patches with neighboring pixels we aggregate features from multiple scales using our proposed spatially aware channel attention mechanism and fuse these features into feature encoding obtained at encoder. 

Motivated by observations of \cite{wang2020high} that highlight domain gap arising due to sensitivities of CNN towards high frequency (HF) components within an image and the ability of adversarial training to incorporate domain invariant properties by focusing upon generalizable patterns within samples. We propose prior based dual discriminators, that use low frequency (LF) and high frequency components along with corresponding dehazed image to determine similarity between recovered image and ground truth. We thus summarize our contributions as \begin{itemize}
    \item Propose an end-to-end dehazing algorithm, directly recovering images affected by unknown haze distribution using spatially aware channel attention mechanism within CNN architecture to ensure feature enhancement and consistency between recovered and neighboring pixels.
    \item Integrate a local augmentation technique, to ensure the network learns to identify and recover haze affected regions in real and synthetic images.
    \item Perform exhaustive experiments to highlight performance inconsistencies between networks trained on synthetic and real datasets and attribute this to weak modeling of haze. 
    \item To ensure consistent performance across synthetic and real datasets, prior based adversarial training mechanism is introduced that leverages LF and HF components within an image to ensure retention of color and structural properties within recovered image.
\end{itemize}

\section{Related Works}
\textbf{Single Image Dehazing :} Image dehazing algorithms can be categorized into model based and end-to-end. Model based algorithms utilize atmospheric scattering model to recover haze affected images either using prior or learning based approach. Prior based approaches such as \cite{he2010single} proposed using dark channel prior on the premise that pixel value within a color channel is close to zero in regions affected by haze, using which transmission map and atmospheric light within an image can be estimated. Other approaches \cite{zhu2015fast, fattal2014dehazing, Berman_2016_CVPR} devise priors such as color attenuation and color lines for estimating transmission map. Since prior based methods are sensitive towards environment variations, learning based approaches are utilized to leverage feature extraction capabilities of CNNs to estimate different components of atmospheric scattering model. Specifically, \cite{lu2016single} used CNNs to estimate atmospheric light,  \cite{cai2016dehazenet} estimates transmission and \cite{yang2017image, li2018cascaded} estimates both transmission and atmospheric light to recover regions affected by haze.

\begin{figure*}[!t]
\centering
\includegraphics[width=\textwidth, height=6.5cm]{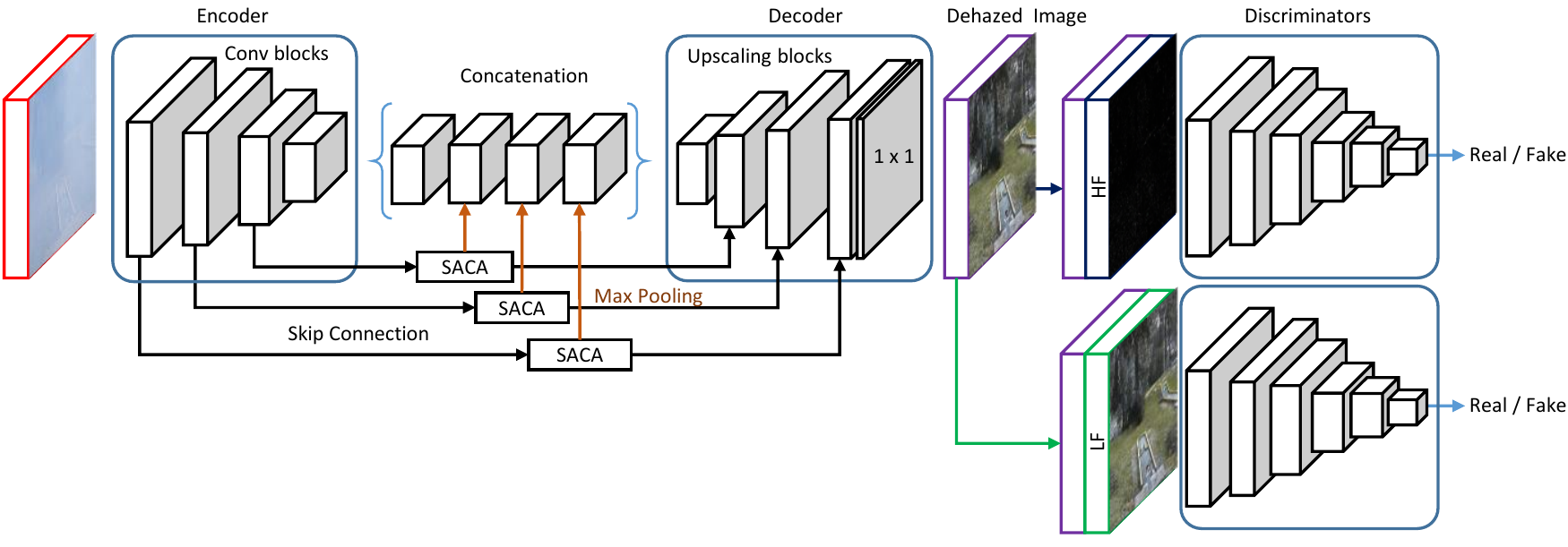}
\caption{An overview of the complete framework covering dehazing and adversarial model}
\label{fig:fig_2}
\end{figure*}

Recently learning based approaches have shown considerable performance improvement for recovering haze affected regions in an end-to-end manner. \cite{ren2018gated} proposed an encoder-decoder formulation to encode features from the hazy images which are then extrapolated by a decoder to reconstruct haze free images. \cite{qin2020ffa, mei2018progressive, li2018cascaded, DuRN_cvpr19} followed a similar approach by performing modifications in CNN network and loss functions.  On contrary \cite{qu2019enhanced} proposed the task of dehazing as image-to-image translation and used a modified variant of Pix2Pix \cite{isola2017image}. To reduce reliance on paired dataset, \cite{engin2018cycle} modified CycleGAN \cite{CycleGAN2017} formulation for the task of dehazing. However these approaches are sensitive towards domain changes between synthetic and real datasets. To overcome this \cite{shao2020domain} proposed a domain adaptation mechanism to translate image from one domain to another thereby aiming to achieve the best of image translation and dehazing. In order generate more visually pleasing dehazed images, \cite{dong2020fd} proposed a fusion of frequency priors with image in adversarial learning framework. However, unlike prior approaches we emphasize on disentangling frequency information (LF and HF) from an image to extract multiple priors and independently learn association between LF (color) and HF (edge) components for retaining color and structural consistency, beyond the traditional loss functions.

\textbf{Domain Invariance :} Feature extraction capabilities of CNNs leads to their SoTA performance on various tasks. However performance inconsistencies arise when domain gap exists between test and train sets. To overcome such scenarios domain adaptation is proposed to perform either feature level \cite{tsai2018learning, tzeng2017adversarial} or pixel level adaptation \cite{shrivastava2017learning, bousmalis2017unsupervised, dundar2018domain, hoffman2018cycada}. Feature level adaptation minimizes maximum mean discrepancy \cite{long2015learning} between source and target domain, while pixel level adaptation focuses upon image-to-image translation or style transfer to increase data in source or target. However reliance on target dataset makes this approach less favorable. An extension, domain generalization, focuses on techniques that provide consistent performance across unknown domains by emphasizing on the task using stylization techniques \cite{somavarapu2020frustratingly, matsuura2020domain}, semantic features \cite{dou2019domain} or adversarial training \cite{li2018domain, agarwal2020intriguing}. Using image stylization or semantic features is not possible since the input images are already affected by haze thereby obscuring underlying information. Thus we focus upon utilizing adversarial learning to achieve domain invariant performance as this method doesnt modify input image. We further lend support from observations by \cite{wang2020high} on importance of LF and HF components within an image, where LF component capture structure and color information while the HF components capture edge information and effectively learning both results in achieving domain invariant performance.

\section{Achieving Domain Invariant Dehazing}
\begin{figure}[ht]
    \centering
    \renewcommand{\tabcolsep}{1pt} 
    \renewcommand{\arraystretch}{1} 
    \begin{adjustbox}{width=0.98\columnwidth}
    \begin{tabular}{ccc}
    \includegraphics[width=0.33\columnwidth, height = 2.5cm]{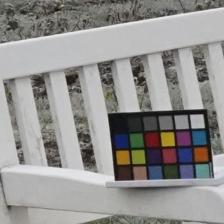} &
    \includegraphics[width=0.33\columnwidth, height = 2.5cm]{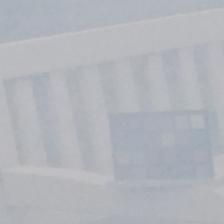} &
    \includegraphics[width=0.33\columnwidth, height = 2.5cm]{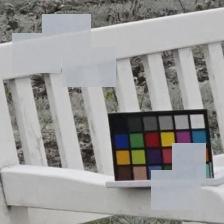} \\
    (a) & (b) & (c) \\
    \includegraphics[width=0.33\columnwidth, height = 2.5cm]{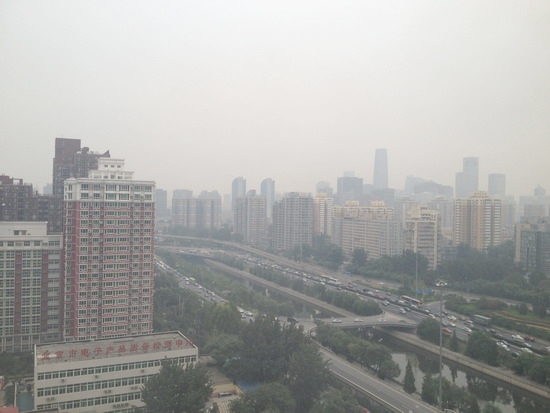} &
    \includegraphics[width=0.33\columnwidth, height = 2.5cm]{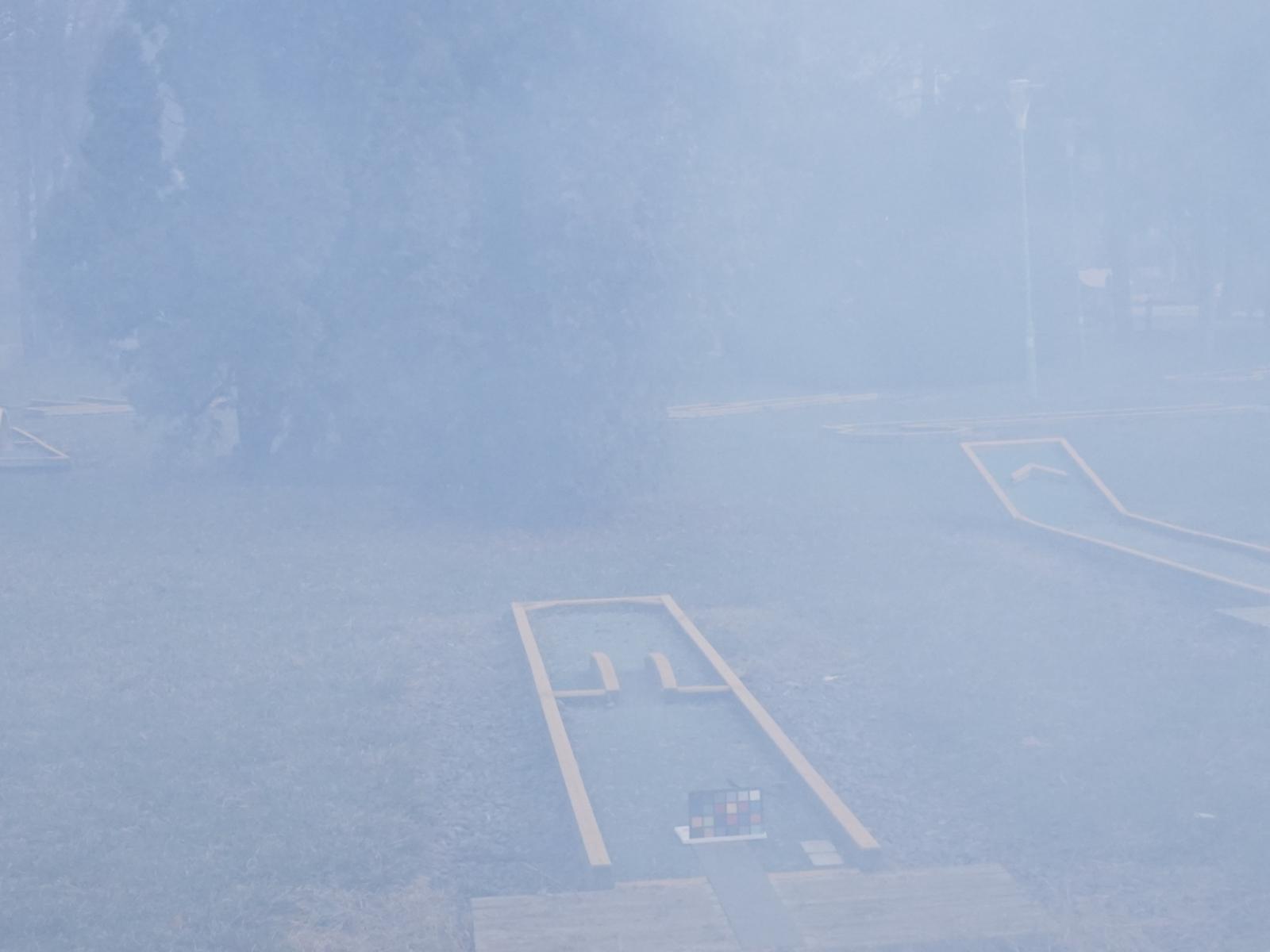} &
    \includegraphics[width=0.33\columnwidth, height = 2.5cm]{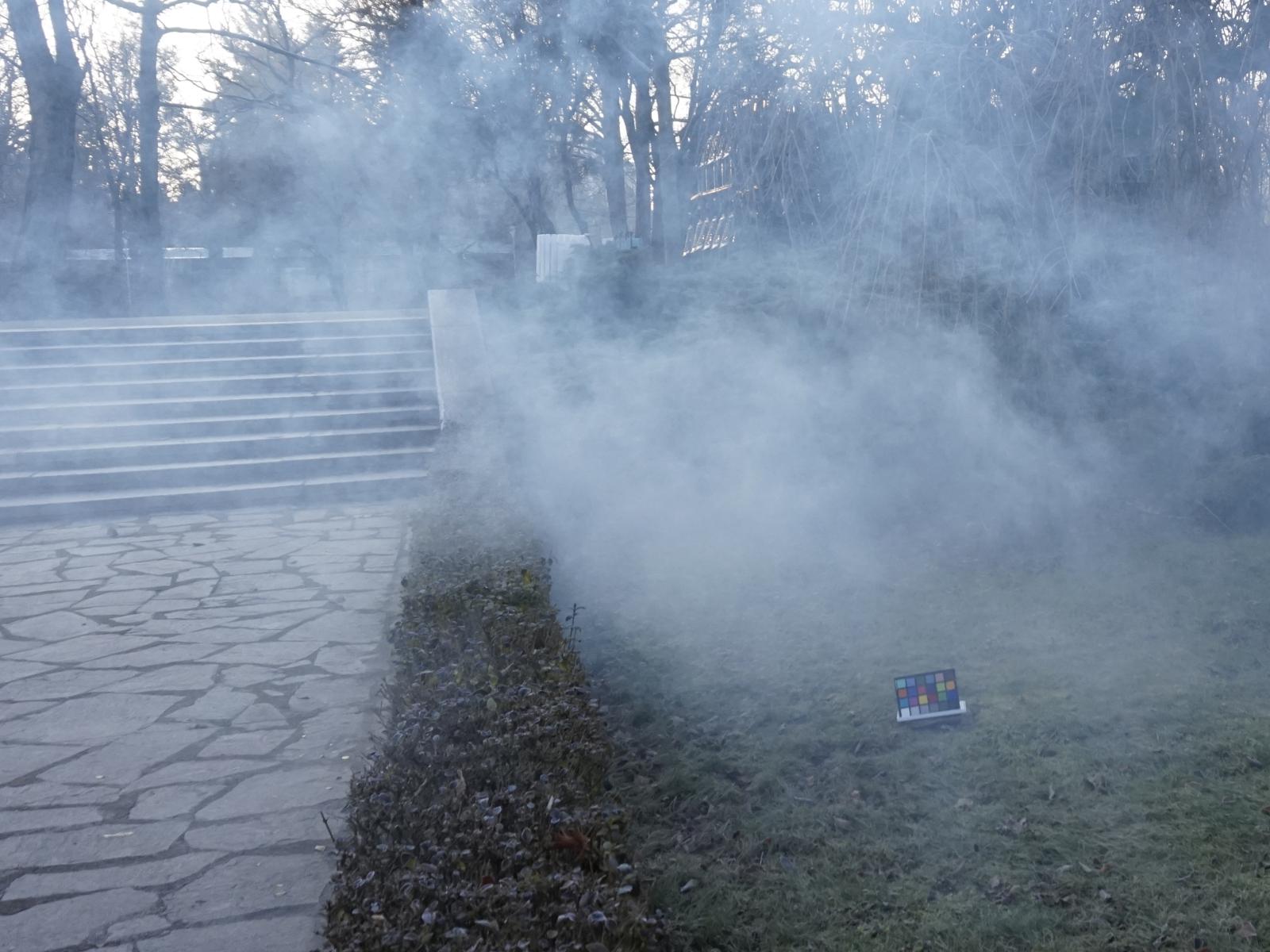} \\
    (d) & (e) & (f) \\
    
    \end{tabular}
    \end{adjustbox}
    \caption{Given a patch of (a) a clean and (b) corresponding paired hazy image to generate (c) localized haze affected regions with maximum patch size set to 50 $\times$ 50. Samples showcasing (d) synthetic, (e) real homogeneous and (f) real non-homogeneous haze from SOTS-OUT, NTIRE-19 and NTIRE-20 datasets.}
    \label{fig:fig3}
\end{figure}

The overall structure of the proposed framework comprises of two parts namely greedy data augmentation (Fig. \ref{fig:fig3}) and adversarial training framework (Fig. \ref{fig:fig_2}) comprising the dehazing and discriminator network.

\subsection{Greedy Localized Data Augmentation}
In a realistic setting, haze can vary wildly across regions within an image (Fig. \ref{fig:fig3} (f)). However synthetic datasets while providing a large number of paired samples arent able to account for such non-homogeneous variations (Fig. \ref{fig:fig3} (d)), leading to inaccurate recovery in such scenarios (Fig. \ref{fig:fig_5}). A simplistic approach to overcome such a scenario would be to increase the dataset to account for these variations which is costly and time consuming. Thus to train the network towards such diverse scenarios, we leverage greedy localized data augmentation technique proposed in \cite{shyam2020copy} to generate small hazy patches of random shape and size within clean images and task the CNN to recover these affected images. This allows utilization of both real and synthetic datasets for homogeneous and non-homogeneous dehazing. Fig. \ref{fig:fig3} demonstrates the working of this augmentation mechanism, along with a visual comparison to synthetic, real and non-homogeneous haze examples. 

\subsection{Network Architecture}
For recovering haze affected regions we build upon encoder-decoder architecture UNet. The encoder of the proposed dehazing network is tasked to represent noisy input image into a latent space using 4 convolution blocks (conv-blocks) that extract relevant features across different scales. Each conv-block represents 2 chains of convolutional filter of size 3 $\times$ 3, batch normalization layer and ReLU activation function. A max pool operation is performed after each convolution block to aggregate features while increasing the receptive field size for subsequent convolution blocks. In order to reconstruct noise-free image, 4 upscaling blocks, each comprising of pixel shuffle layer \cite{shi2016real} followed by convolutional filter of size 3 $\times$ 3, batch normalization and ReLU activation are used. Features obtained at each level of encoder are concatenated with features of corresponding level at the decoder end via long skip connections. This ensures presence of fine grained features extracted in early layers to be present in the noise-free image, helping to maintain boundary properties of objects present within image. The result from decoder is passed to a convolution layer with filter size 1 $\times$ 1. 

\subsubsection{Spatially Aware Channel Attention (SACA) :} 
\begin{figure}[ht]
    \centering
    \includegraphics[width=\columnwidth, height = 3.25cm]{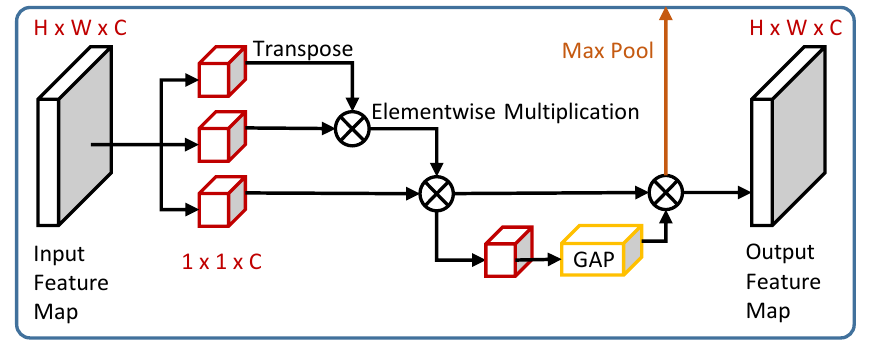} \\
    \caption{Structural overview of proposed Spatially Aware Channel Attention mechanism}
    \label{fig:fig4}
\end{figure}

While such an encoder-decoder architecture tends to work on homogeneous haze distribution. In case of non homogeneous haze, affected regions might extend beyond receptive field of convolutional kernels resulting in weak representation being extracted along different scales of encoder. Thus dynamic adjustment of receptive field based on haze distribution for encompassing relevant features is required. For this we propose a spatially aware channel attention mechanism that comprises of non-local operation \cite{wang2018non} followed by a channel attention mechanism. The non local operation allows in capturing long-range dependencies across spatial dimension, while channel attention mechanism filters important channels within feature map. Adding the channel attention mechanism helps in reducing the computational cost, thus allowing deployment of such blocks at different scales. The channel attention mechanism (Fig. \ref{fig:fig4}) is constructed using 1 $\times$ 1 convolution, global average pooling and a softmax activation layer and works by amplifying relevant channels while suppressing irrelevant ones. In order to maximize the effect of spatially aware channel attention layer, we place them in long skip connections and justify such a design choice for ensuring long skip connections carry relevant local features by refining complete feature map corresponding to a particular scale without modifying features within conv-blocks in encoders. 
The resultant refined features from each SACA block is include into final embedding representation by performing a max pool operation (of varying size) to match feature map size and highlight that such a mechanism allows in enriching the feature space by concatenating additional features.

\begin{figure*}[!ht]
    \centering
    \renewcommand{\tabcolsep}{1pt} 
    \renewcommand{\arraystretch}{1} 
    \begin{adjustbox}{max width=\textwidth}
    \begin{tabular}{cccccc}
    \includegraphics[width=0.166\textwidth, height=2.5cm]{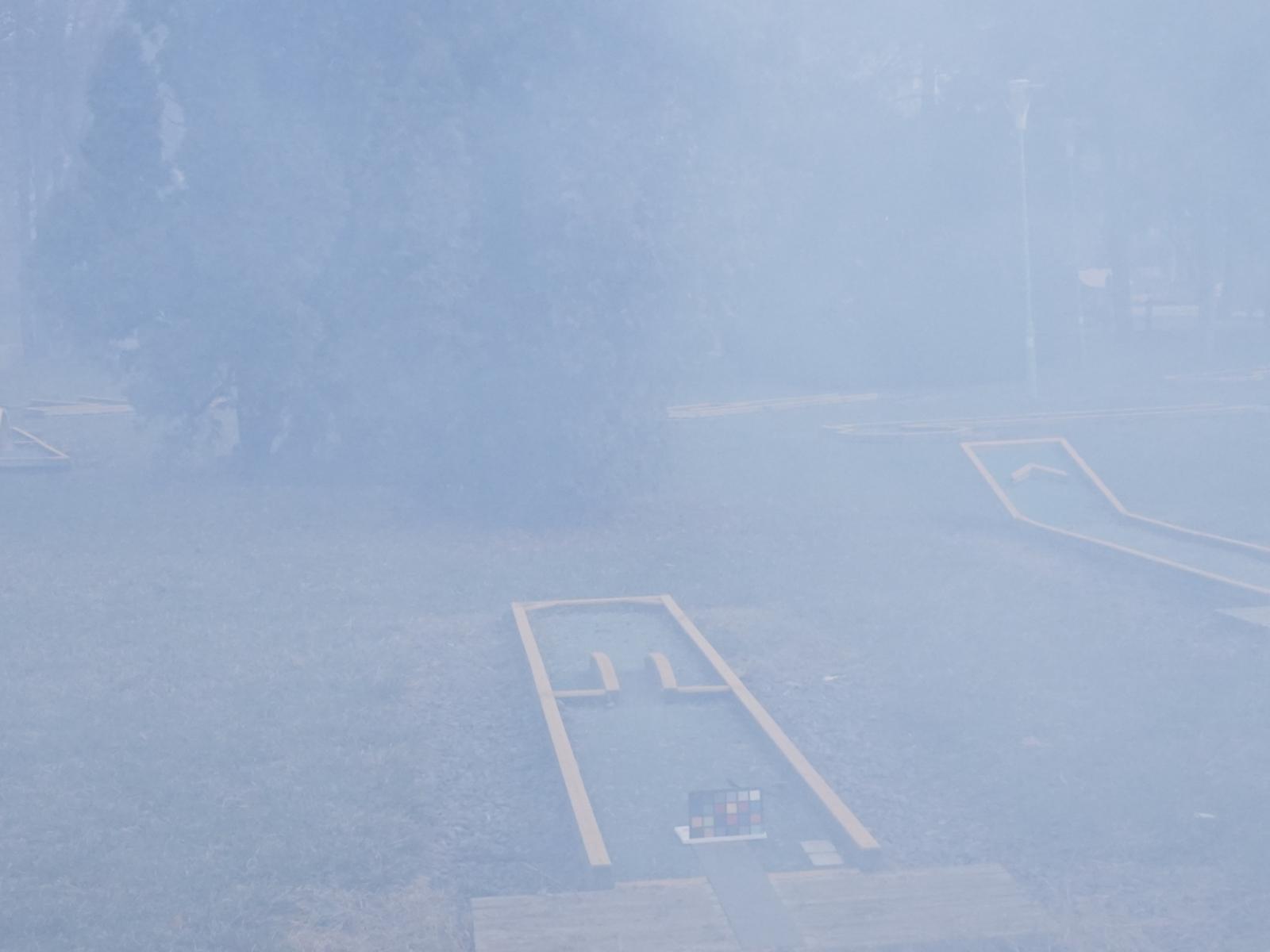} &
    \includegraphics[width=0.166\textwidth, height=2.5cm]{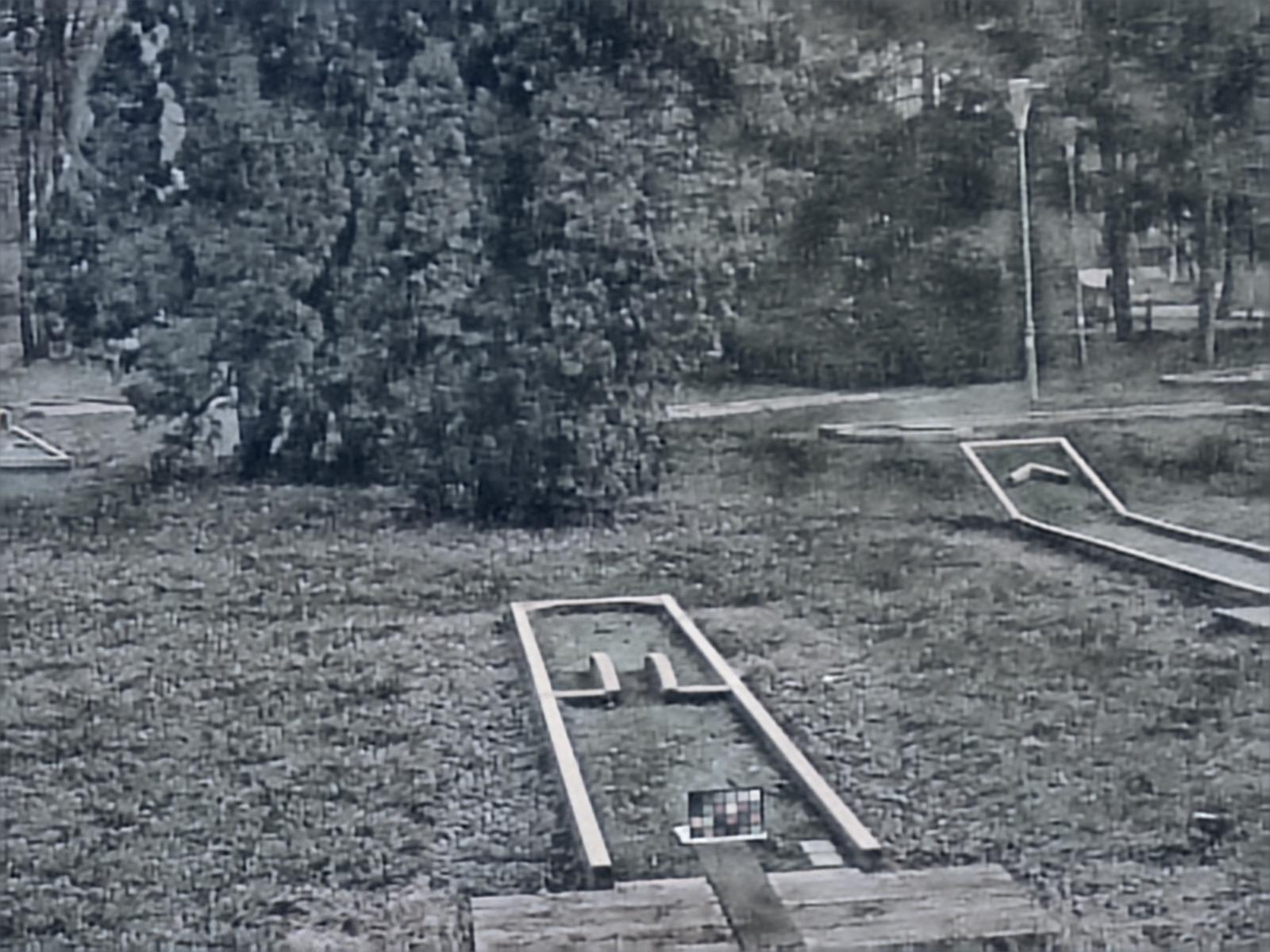} & 
    \includegraphics[width=0.166\textwidth, height=2.5cm]{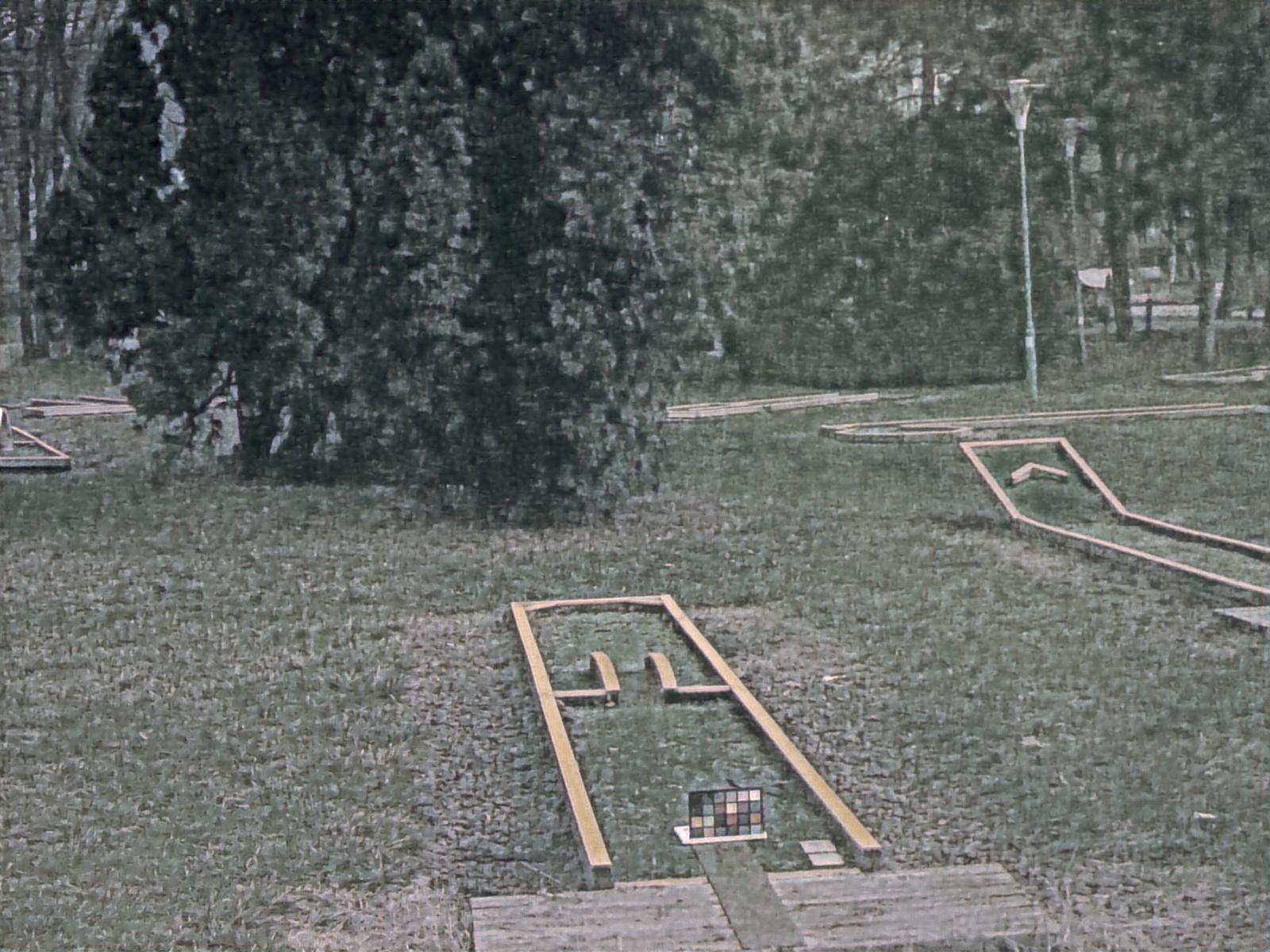} & 
    \includegraphics[width=0.166\textwidth, height=2.5cm]{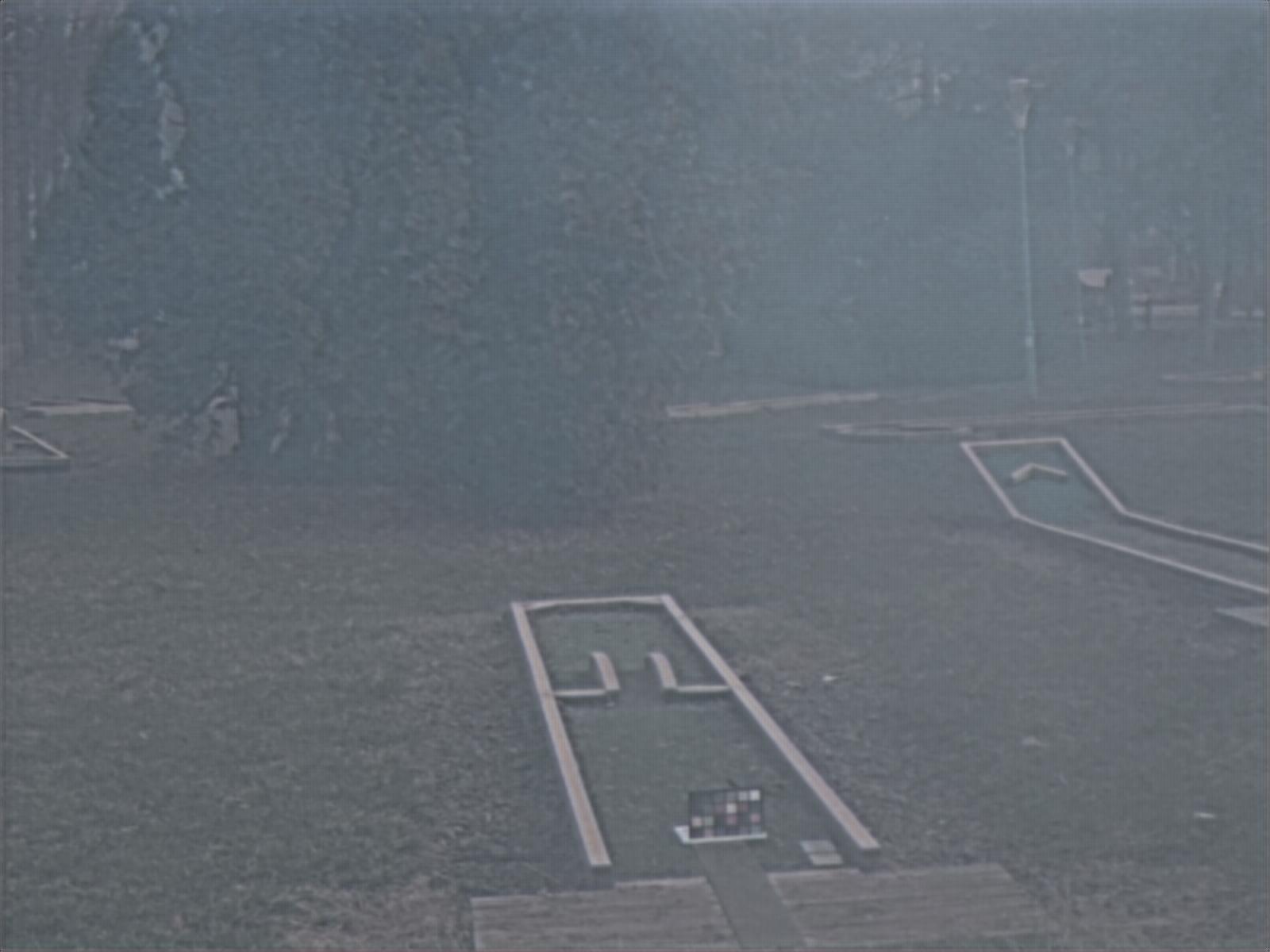} & 
    \includegraphics[width=0.166\textwidth, height=2.5cm]{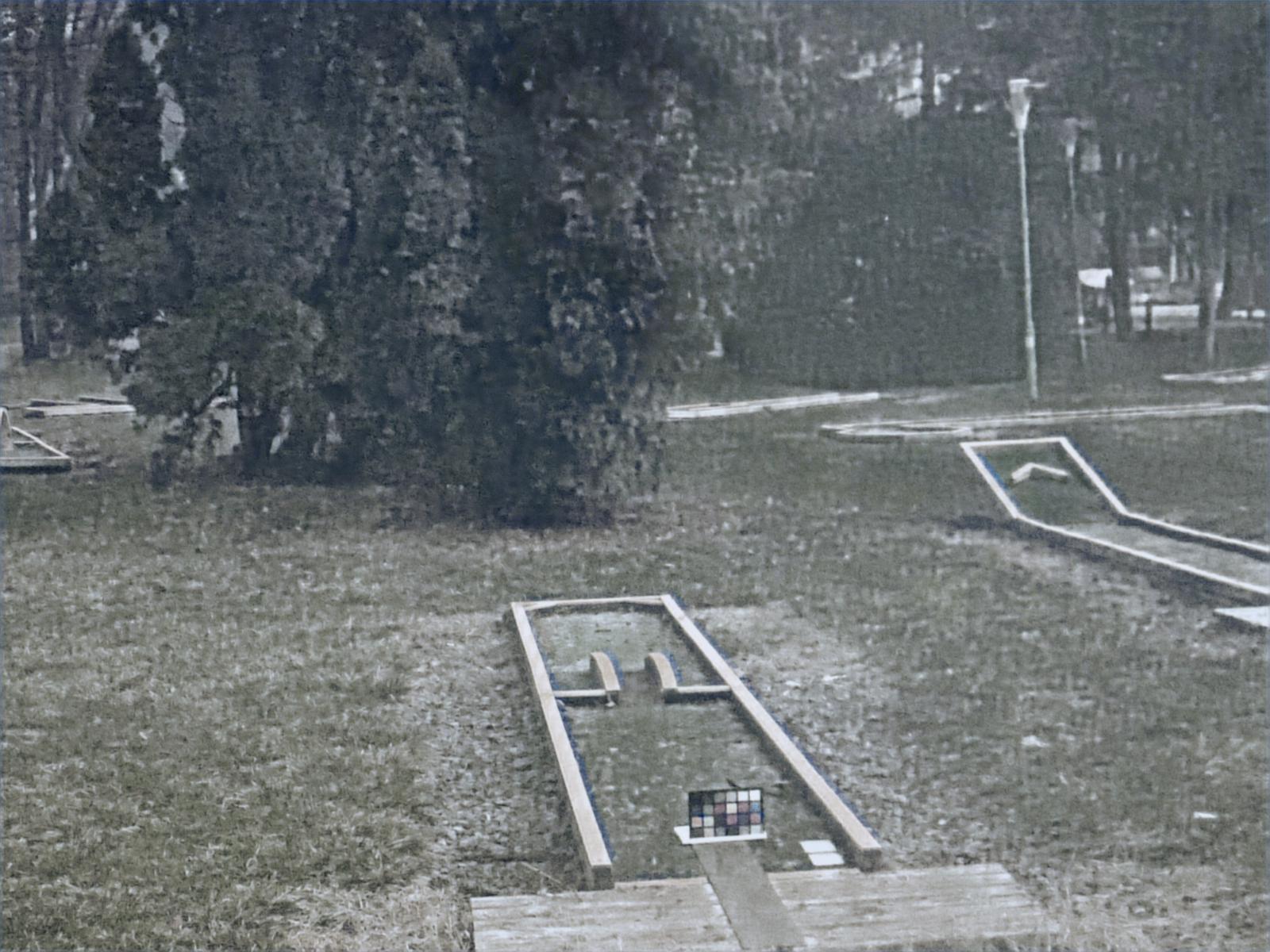} & 
    \includegraphics[width=0.166\textwidth, height=2.5cm]{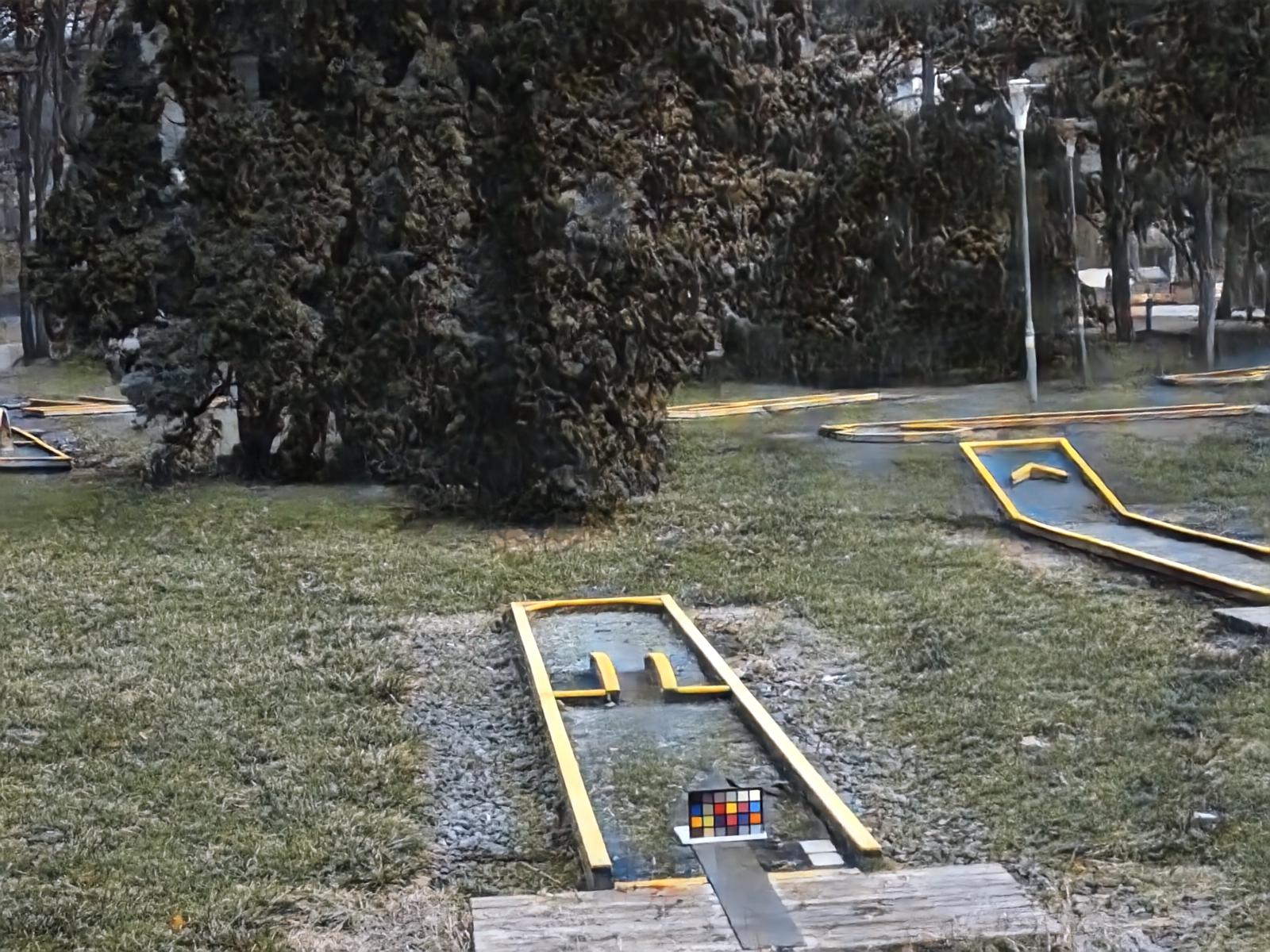} \\ 
    7.09 / 0.33 & 16.76 / 0.55  & 18.18 / 0.62  & 14.07 / 0.43  & 16.39 / 0.60  & 19.07 / 0.58 \\
    Input Image & DuRN-US       & FFA-Net       & Wavelet-UNet  & SNDN          & MSNet  \\    
    \includegraphics[width=0.166\textwidth, height=2.5cm]{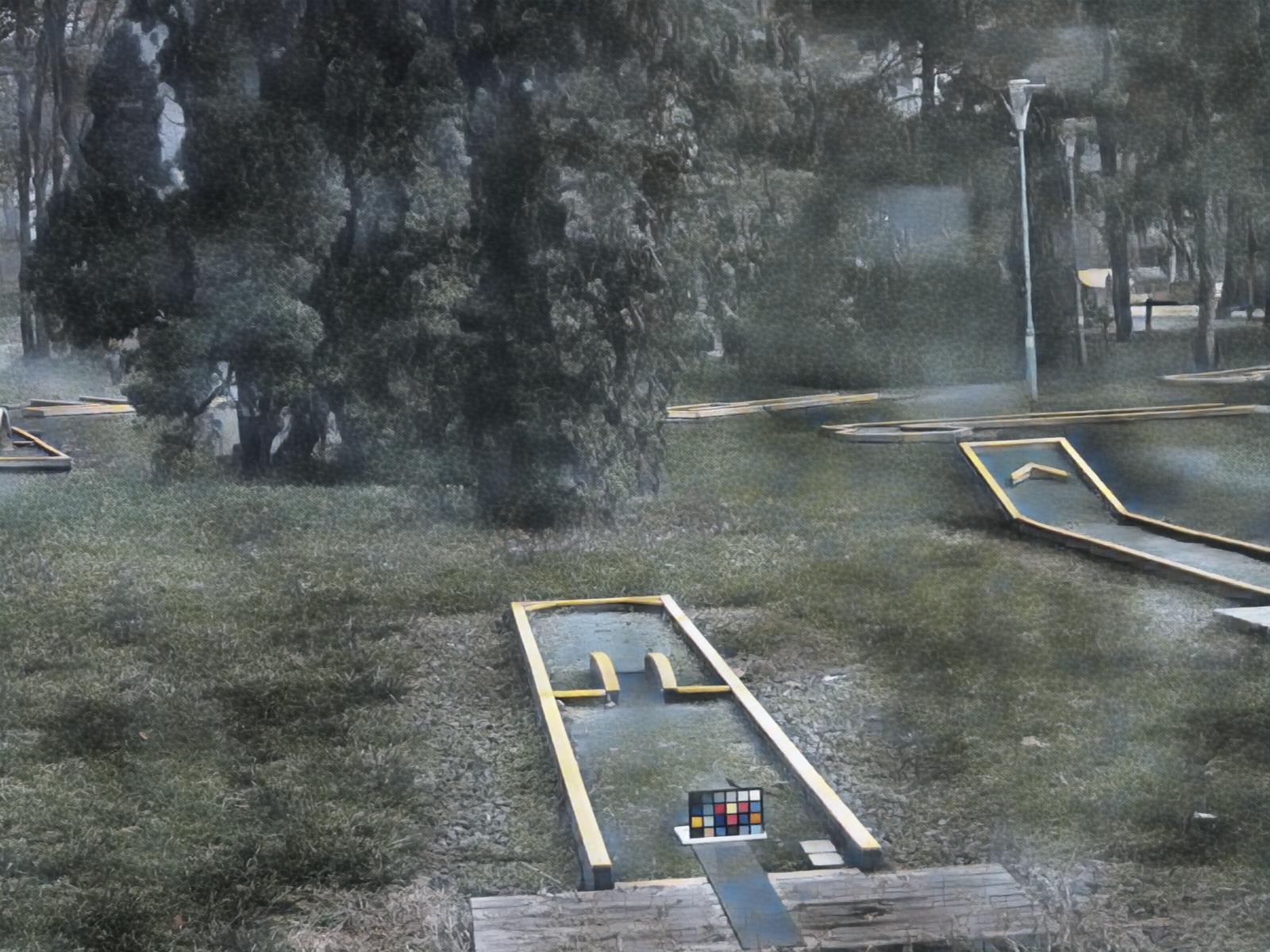} & 
    \includegraphics[width=0.166\textwidth, height=2.5cm]{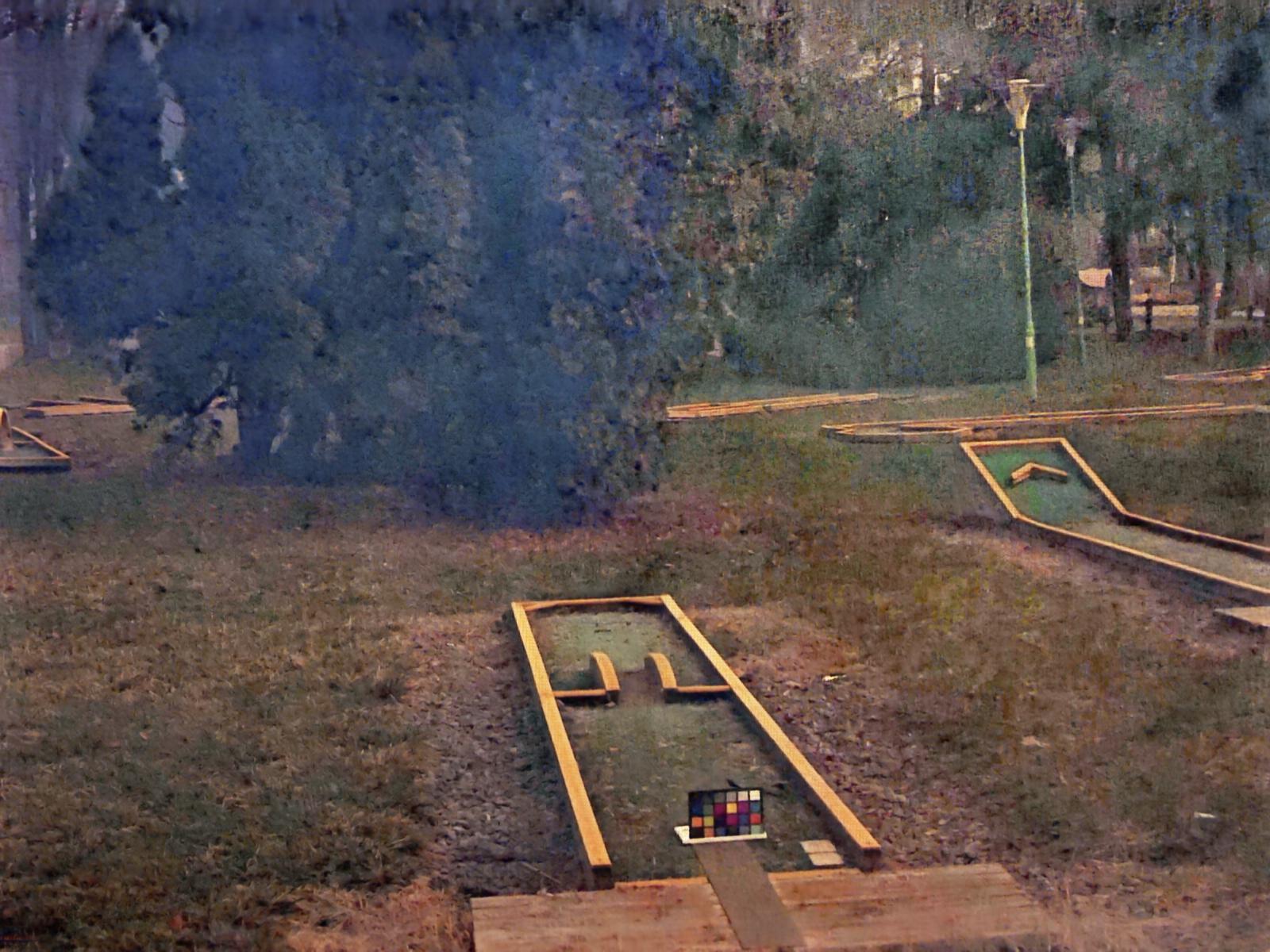} & 
    \includegraphics[width=0.166\textwidth, height=2.5cm]{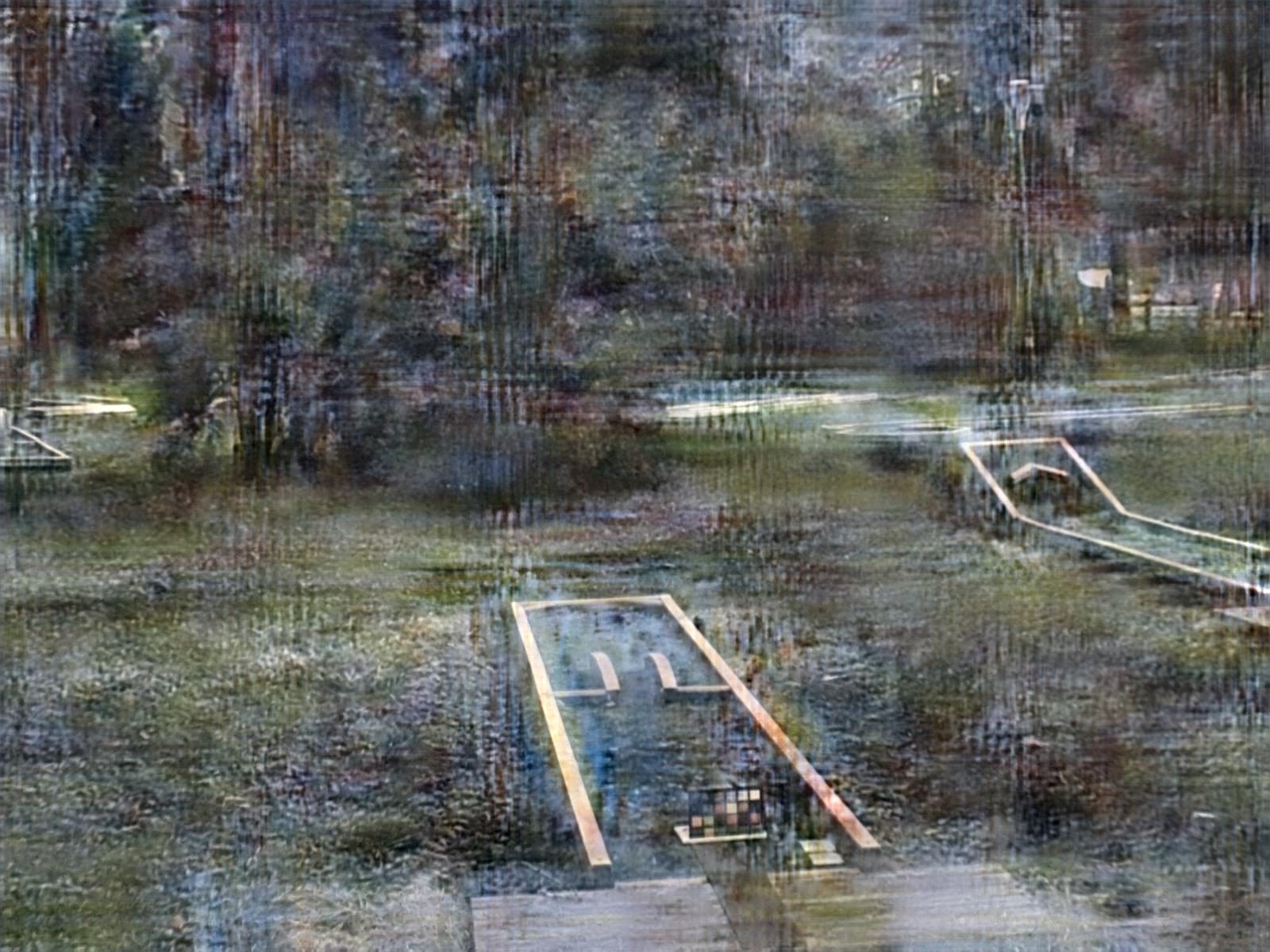} & 
    \includegraphics[width=0.166\textwidth, height=2.5cm]{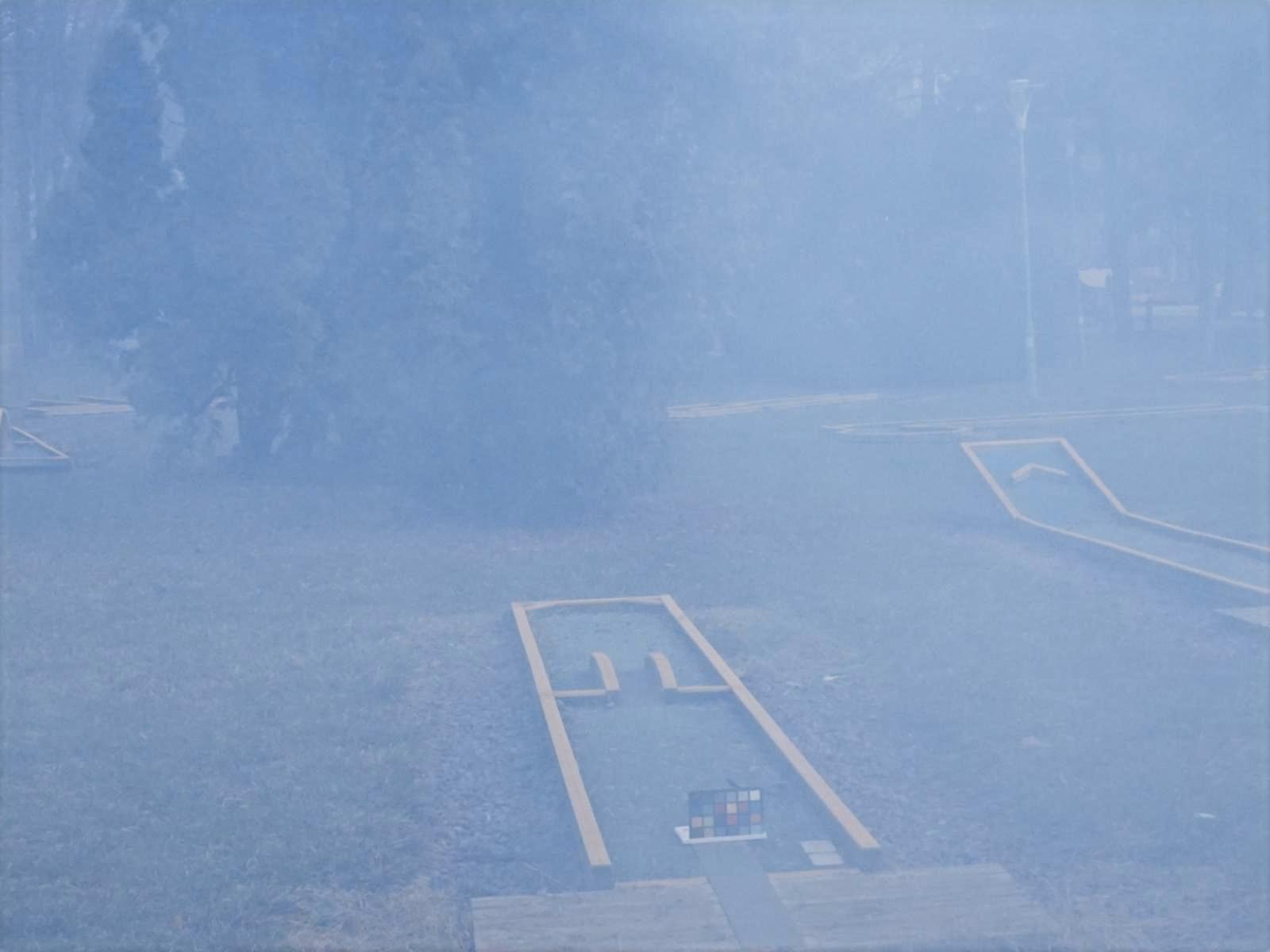} &
    \includegraphics[width=0.166\textwidth, height=2.5cm]{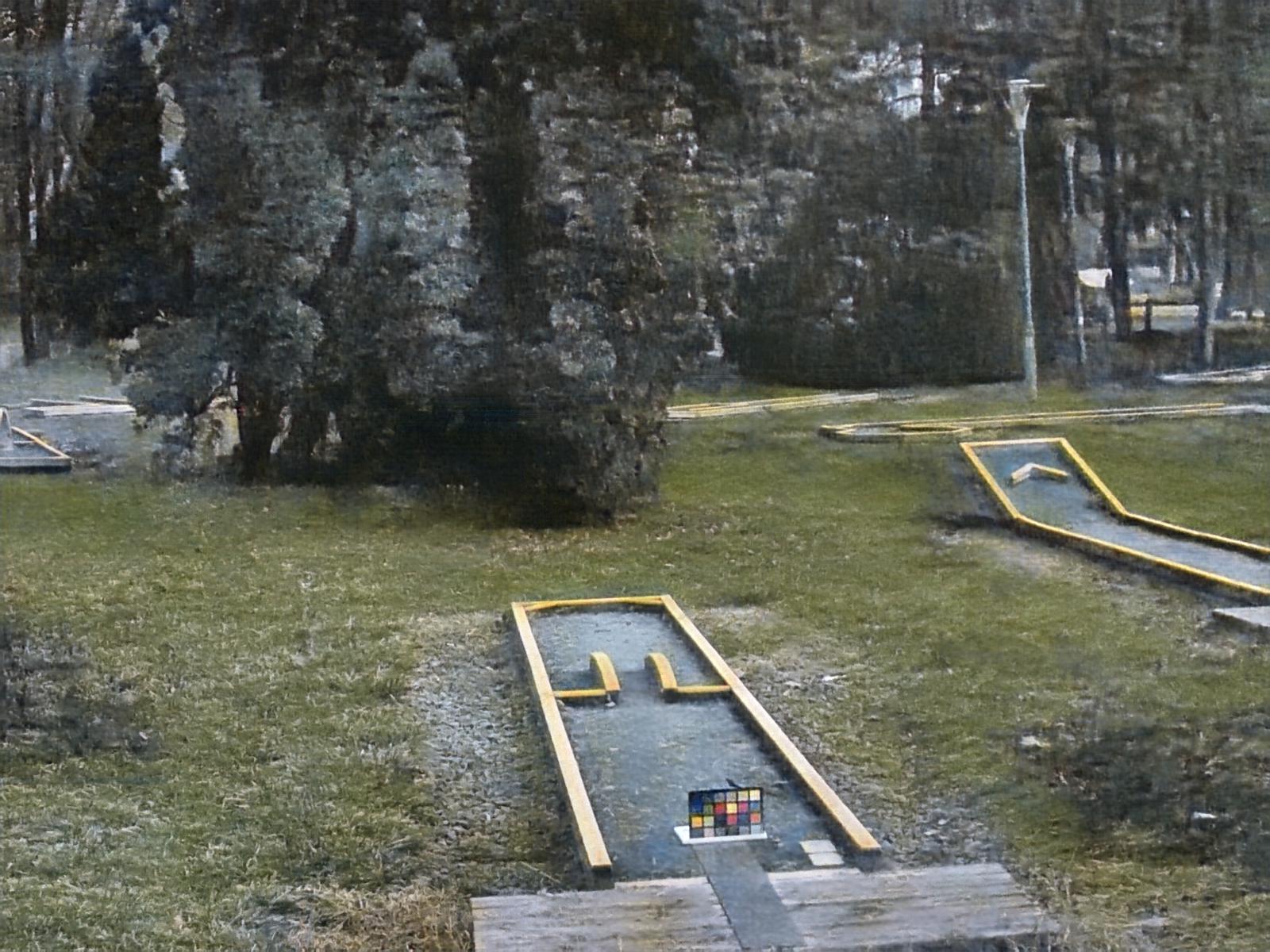} &
    \includegraphics[width=0.166\textwidth, height=2.5cm]{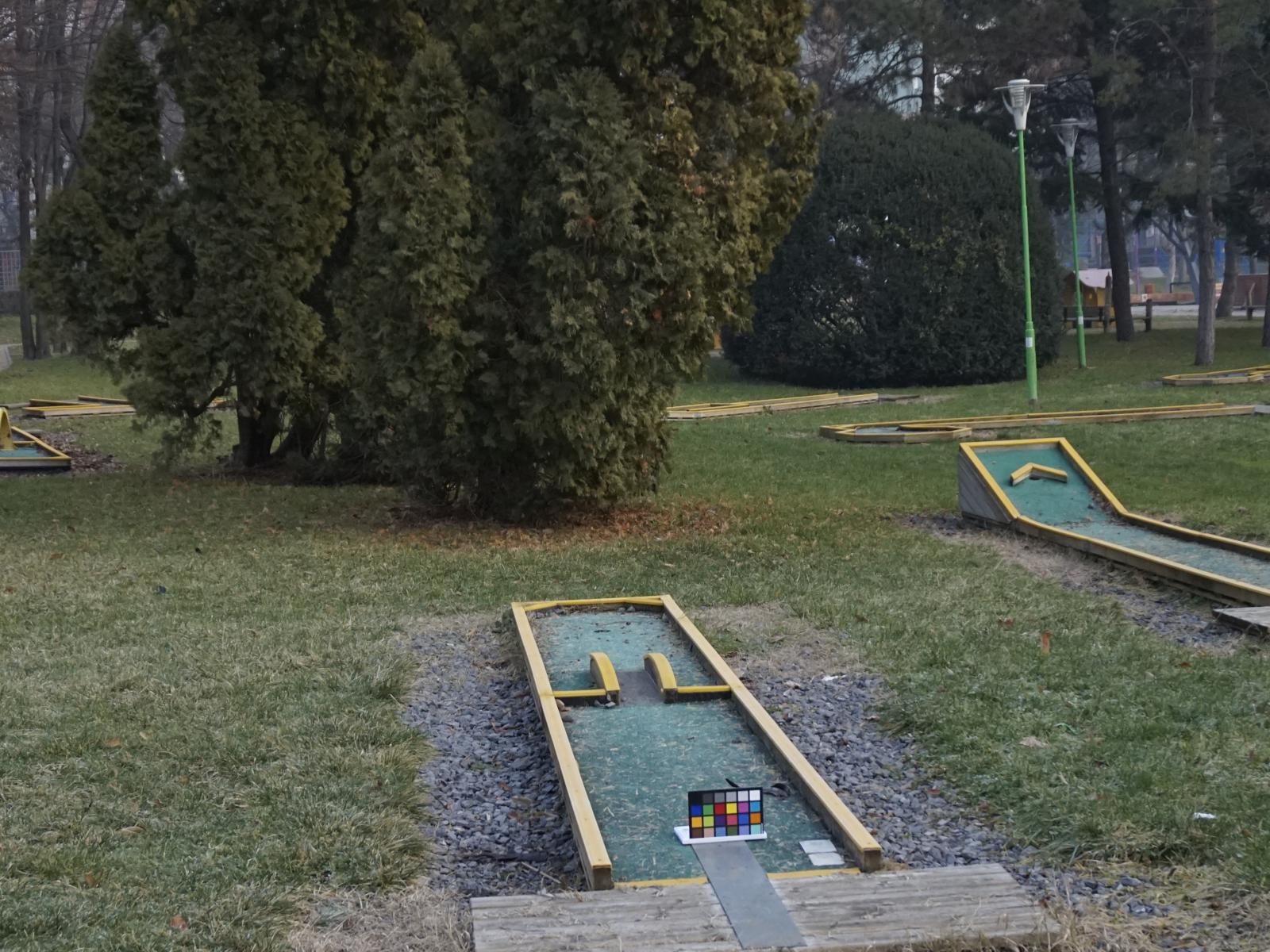} \\    
    17.96 / 0.61  & 17.36 / 0.58 & 16.80 / 0.49& 8.30 / 0.36 & 22.18 / 0.71 & \\
    GridDehazeNet & DA-Dehazing  & PFFNet      & Y-Net       & Ours & Clean Image \\
                
    \end{tabular}
    \end{adjustbox}
    \caption{Performance of different Dehazing algorithms when trained and evaluated on NTIRE-19 dataset.}
    \label{fig:fig_4}
\end{figure*}


\subsection{Frequency prior based Discriminators for Adversarial Training}
Domain gap between synthetic and real haze samples adversely affect performance of underlying dehazing algorithms (Tab. \ref{tab:tab_1}). Thus to obtain a domain invariant dehazing algorithm, we take advantage of frequency domain and propose a frequency prior based discriminator that relies on both high and low frequency components of an image to determine if a recovered image matches ground truth. The discriminator architecture comprises of 6 conv-blocks resulting in multi-dimensional output, similar to Patch-GAN \cite{isola2017image}, with a patch-size of 64. We utilize two independent discriminators with same architecture but using different frequency priors to obtain different set of weights. We base this design choice on two observations, 
\begin{enumerate}
    \item HF components cover edge information, while LF components cover structure and color information. In this context the intensity of LF components would be larger than HF components, which might lead to LF components gaining more importance in adversarial process.
    \item \cite{wang2020high} highlights during early optimization process LF components are learned owing to a steeper descent of loss surface.
\end{enumerate}

These observations incline us to introduce two discriminators to avoid over reliance of one component over other, while optimizing the complete framework. Monitoring the optimization process ascertains that both LF and HF components are learned. To train the discriminators, for a given image, we first extract its high and low frequency components using laplacian and gaussian filters respectively (of filter size 3 and 7 respectively) and concatenate them with original image. To ensure standard pixel scale, we normalize the HF components before concatenation. Thus for a given pair of hazy $I_N$ and its corresponding dehazed image $I_R$, the dehazing network estimates a dehazed image $G(I_N)$. Corresponding LF $(LF(.))$ and HF $(HF(.))$ components are extracted and concatenated, resulting in $([I_R, LF(I_R)], [I_R, HF(I_R)]$ and $([G(I_N), LF(G(I_N))], [G(I_N), HF(G(I_N))]$ prior based samples. These samples are used as inputs for corresponding low and high frequency discriminators, that would classify them as real or fake. Thus while training the discriminators would follow the min-max optimization cycle,

\begin{equation}
\begin{split}
    \underset{G}{\min} \ \underset{D_{LF},\ D_{RF}}{\max} \quad &  \underset{I_R \sim \ \mathbb{R}_{real}} {\mathbb{E}} \ \{ log (D_{LF} \ [I_R,\ LF(I_R)]) \\
    & + \ log (D_{LF} \ [I_R,\ HF(I_R)]) \}  \\ \\
    + \underset{G(I_N) \sim \ \mathbb{R}_{fake}}{\mathbb{E}} & \{ log (1 - D_{LF} [G(I_N),\ LF(G(I_N))]) \\
    + & \ log (1 - D_{HF} [G(I_N),\ HF(G(I_N))]) \} \\
\end{split}    
\end{equation}

While FFTs can be computed individually on each separate channel of an image or gray scale image there exists another approach namely on channe

\subsection{Framework Optimization}
To train the proposed framework, we follow standard GAN approach wherein the dehazing algorithm and discriminator are optimized alternatively. The optimization function for the dehazing algorithm is composed of L1, SSIM \cite{wang2004image} and perceptual \cite{johnson2016perceptual} losses along with dual adversarial loss. 

\begin{equation}
\begin{split}
L_{G}(I_R, & \ G(I_N)) = L_1(I_R, G(I_N)) + SSIM(I_R, G(I_N)) \\
    & + L_{VGG}(I_R, G(I_N)) \\
    & + \lambda_1*log (1 - D_{LF} [G(I_N),\ LF(G(I_N))]) \\
    & + \lambda_2*log (1 - D_{HF} [G(I_N),\ HF(G(I_N))])
\end{split} 
\end{equation}

\noindent
where $\lambda_1, \ \lambda_2$ are loss balancing terms. In our experiments we set $\lambda_1=\lambda_2=0.5$ to balance both LF and HF discriminators. We design the proposed framework in Pytorch 1.6. The input patch is set to square patches of size 512 normalized to $[0, 1]$. ADAM \cite{kingma2014adam} is used as optimizer with $\beta_1=0.5$ and $\beta_2=0.9$ and learning rate of 0.0001 for dehazing and 0.0003 for discriminator networks respectively with a batch size of 4. Apart from the aforementioned greedy localized data augmentation (max patch size of $50 \times 50$), we also use random horizontal and vertical flipping as additional augmentation techniques. For our experiments we utilize a system equipped with Intel 8700-K CPU and 64GB RAM having Nvidia Titan V and Titan RTX GPUs.

\section{Experimental Evaluations}

\textbf{Datasets and Evaluation Metrics :} In order to evaluate performance of various algorithms across both synthetic and real datasets, exhibiting different haze distributions. We utilize real i.e. NTIRE-18 \cite{ancuti2018ntire}, NTIRE-19 \cite{cai2019ntire}, NTIRE-20 \cite{yuan2020ntire} and synthetic i.e. SOTS \cite{li2019benchmarking} and Haze-RD \cite{Zhang} datasets and summarize their properties such as resolution, haze type and average PSNR and SSIM of hazy images in Tab. \ref{tab:tab_1}. For evaluating the NTIRE-20 dataset, we first create a subset from training sample and utilize the remaining dataset for training. Furthermore to compare the performance of different SoTA algorithms we utilize Peak-Signal-to-Noise Ratio (PSNR) and Structural Similarity Metric (SSIM) as evaluation metrics. We utilize open-source CNN based dehazing methods such as DuRN-US \cite{DuRN_cvpr19}, FFA-Net \cite{qin2020ffa}, Wavelet-UNet \cite{yang2019wavelet}, SNDN \cite{chen2020simplified}, MSNet \cite{msnet2020}, GridDehazenet \cite{liuICCV2019GridDehazeNet}, PFFNet \cite{mei2018pffn}, AtJ-DH+ \cite{guo2019dense}, KTDN \cite{ancuti2020ntire}, TridentNet \cite{liu2020trident}, Enhanced-Pix2Pix \cite{qu2019enhanced} and DA-Dehaze \cite{shao2020domain}.

\begin{table}[!h]
    \caption{Properties of Different Datasets} 
    \begin{adjustbox}{max width=\columnwidth}
    \begin{tabular}{lccc}
    \Xhline{2\arrayrulewidth} \hline \noalign{\vskip 1pt}
    Dataset Name    &  PSNR / SSIM & Resolution         & Type \\ 
    \hline \noalign{\vskip 3pt}
    NTIRE-18        & 14.60 / 0.67 & 4177 $\times$ 3134 & Real \\
    NTIRE-19        &  9.11 / 0.49 & 1600 $\times$ 1200 & Real \\
    NTIRE-20        & 10.42 / 0.46 & 1600 $\times$ 1200 & Real \\
    SOTS-IN         & 11.97 / 0.69 & 620 $\times$ 460   & Synthetic \\
    SOTS-OUT        & 15.92 / 0.81 & 550 $\times$ 478   & Synthetic \\
    HazeRD          & 14.60 / 0.67 & 3492 $\times$ 2558 & Synthetic \\
    \Xhline{2\arrayrulewidth} \hline
    \end{tabular}
    \end{adjustbox}
    \label{tab:tab_1}
\end{table}

\begin{table*}[!ht]
    \centering
    \caption{Model Performance when trained independently on synthetic and real datasets}
    \begin{adjustbox}{max width=\textwidth}
    \begin{tabular}{lcccc||cccc}
    \Xhline{3\arrayrulewidth} \hline \noalign{\vskip 1pt}
    & \multicolumn{4}{c||}{Trained on Synthetic (RESIDE-Indoor) dataset} & \multicolumn{4}{c}{Trained on Synthetic (RESIDE-Outdoor) dataset} \\
    \cline{2-9} \noalign{\vskip 1pt}
    & SOTS-IN & SOTS-Out & NTIRE-19 & NTIRE-20 & SOTS-IN & SOTS-Out & NTIRE-19 & NTIRE-20 \\
    \Xhline{2\arrayrulewidth} \hline \noalign{\vskip 1pt}
    DuRN-US         & 32.12 / \textbf{0.98} & 19.55 / 0.83 & 10.81 / \underline{0.51} & \underline{11.27} / \underline{0.51} & 15.95 / 0.76 & 19.41 / 0.81 & 11.04 / 0.51 & 11.73 / 0.46 \\ 
    FFA-Net         & \underline{36.36} / \textbf{0.98} & 20.05 / \underline{0.84} & 10.97 / 0.42 & 10.70 / 0.44 & 18.96 / 0.86 & 30.88 / 0.93 & 09.64 / 0.50 & 10.90 / 0.48 \\
    Wavelet-UNet    & 20.02 / 0.75 & 17.75 / 0.67 & \underline{11.48} / 0.47 & 10.88 / 0.36 & 16.26 / 0.73 & 21.95 / 0.76 & 10.36 / 0.49 & 11.05 / 0.43 \\
    MSNet           & 32.04 / \textbf{0.98} & \underline{20.70} / \textbf{0.86} & 09.90 / \underline{0.51} & 11.16 / \underline{0.51} & 21.75 / 0.88 & 29.80 / 0.93 & 09.56 / 0.48 & 11.35 / 0.51 \\
    SNDN            & 24.68 / 0.91 & 16.02 / 0.69 & 10.13 / 0.45 & 10.64 / 0.43 & 25.30 / 0.91 & 24.31 / 0.88 & 11.74 / 0.49 & 11.95 / 0.52 \\
    GridDehazeNet   & 32.14 / \textbf{0.98} & 16.22 / 0.76 & 09.50 / 0.49 & 09.01 / 0.40 & 20.99 / 0.89 & 29.18 / 0.93 & 10.16 / 0.50 & 11.23 / 0.49 \\
    PFFNet          & 26.58 / \underline{0.92} & 14.63 / 0.65 & 11.38 / \underline{0.51} & 11.14 / 0.43 & 20.32 / 0.85 & 27.65 / 0.91 & 10.75 / 0.50 & 11.55 / 0.52 \\
    Enhanced-Pix2Pix&  25.06 / 0.92 & - & - & - & - & 22.57 / 0.86 & - & - \\
    Ours            & \textbf{38.91} / \textbf{0.98} & \textbf{25.75} / \underline{0.84} & \textbf{16.21} / \textbf{0.78} & \textbf{16.28} / \textbf{0.67} & 26.90 / 0.76 & 30.40 / 0.94 & 13.36 / 0.52 & 12.68 / 0.52 \\
    
    \Xcline{2-9}{2\arrayrulewidth} \cline{2-9} \noalign{\vskip 1pt}
    & \multicolumn{4}{c||}{Trained on Real (NTIRE-19) dataset} & \multicolumn{4}{c}{Trained on Real (NTIRE-20) dataset} \\
    \cline{2-9} \noalign{\vskip 1pt}
    & SOTS-IN & SOTS-Out & NTIRE-19 & NTIRE-20 & SOTS-IN & SOTS-Out & NTIRE-19 & NTIRE-20 \\
    \Xcline{2-9}{2\arrayrulewidth} \cline{2-9} \noalign{\vskip 1pt}
    DuRN-US         & 11.44 / \underline{0.59} & 13.05 / 0.61 & 13.63 / \underline{0.57} & 12.97 / 0.52 & 09.43 / 0.63 & 11.92 / 0.66 & 11.63 / \underline{0.52} & 15.27 / 0.50 \\
    FFA-Net         & 12.16 / 0.55 & 14.36 / 0.59 & 14.01 / 0.56 & 14.71 / \underline{0.57} & 09.96 / 0.63 & \underline{14.88} / \underline{0.75} & 12.43 / \underline{0.52} & 18.11 / 0.66 \\
    Wavelet-UNet    & 13.57 / 0.41 & 13.05 / 0.44 & 12.85 / 0.39 & 12.08 / 0.24 & \underline{12.04} / 0.32 & 13.85 / 0.41 & 11.46 / 0.28 & 12.08 / 0.21 \\
    MSNet           & 13.33 / 0.55 & 13.85 / 0.56 & 13.32 / 0.53 & 12.63 / 0.32 & 09.16 / 0.51 & 10.66 / 0.56 & 12.04 / 0.50 & 14.06 / 0.50 \\
    SNDN            & 12.56 / \textbf{0.66} & 14.11 / \underline{0.70} & 13.54 / 0.54 & \underline{14.93} / 0.51 & 12.03 / \underline{0.67} & 14.14 / 0.73 & 11.73 / \underline{0.52} & 13.93 / 0.52 \\
    GridDehazeNet   & 14.57 / \underline{0.59} & 13.47 / 0.60 & 12.96 / 0.50 & 12.07 / 0.32 & 11.60 / 0.58 & 12.75 / 0.72 & \underline{13.39} / \underline{0.52} & 15.32 / 0.60 \\
    PFFNet          & 13.51 / 0.50 & \underline{14.57} / 0.53 & 13.29 / 0.52 & 12.99 / 0.31 & 08.82 / 0.47 & 12.00 / 0.53 & 11.54 / 0.49 & 14.50 / 0.36 \\
    AtJ-DH+         &       -      &       -      & \underline{17.18} / 0.53 &       -      &       -      &       -      &       -      &       -      \\
    KTDN            &       -      &       -      &       -      &       -      &       -      &       -      &       -      & 20.85 / 0.69 \\
    TridentNet      &       -      &       -      &       -      &       -      &       -      &       -      &       -      & \textbf{21.41} / \underline{0.71} \\
    Ours            & \textbf{19.28} / \textbf{0.66} & \textbf{18.17} / \textbf{0.87} & \textbf{19.47} / \textbf{0.75} & \textbf{20.33} / \textbf{0.77} & \textbf{19.53} / \textbf{0.71} & \textbf{18.69} / \textbf{0.79} & \textbf{17.24} / \textbf{0.66} & \underline{21.17} / \textbf{0.78} \\
    \Xhline{3\arrayrulewidth} \hline
    \end{tabular}
    \end{adjustbox}
    \begin{tablenotes}
    \scriptsize
    \item \textbf{Boldface} and \underline{Underlined} values represent best and second best results for each unique independent training dataset. \\
    \item - Indicates missing values not reported in Original Paper
    \end{tablenotes}
    \label{tab:tab_3}
\end{table*}


\textbf{Individual vs Aggregated Dataset for Strong Baseline :} Synthetic datasets provide access to extremely diverse characteristics such as scene setting, differing camera properties and illumination conditions, covered in large amount of paired datasets, making them indispensable despite their flaws in modeling haze. We begin by determining performance of algorithms when trained and evaluated on datasets having same distribution while summarizing the results in Tab. \ref{tab:tab_3}. We observe dehazing algorithms to perform well on test sets following similar distribution to training dataset, but their performance drops drastically when tested on datasets outside the training distribution, even for algorithms trained on synthetic samples. However, we observe that compared to previous methods, the performance drop of proposed approach is not substantial and we attribute this to utilizing frequency priors while training.

A common approach to achieve domain invariant performance is to increase the dataset size by accumulating data from different sources. Following this we aggregate the aforementioned datasets and evaluate them individual sub test sets with results summarized in Tab. \ref{tab:tab_2}. We conclude this approach to aid in achieving peak performance for all algorithms on real datasets such as NTIRE-19 and NTIRE-20. We further corroborate that all algorithms including ours benefit from increased dataset size on account of merging synthetic and real datasets. In this scenario the proposed model outperforms the top algorithm DuRN-US by 3.78 db, 0.19 and 6.84db and 0.21 on NTIRE-20 and NTIRE-19 datasets, thus showcasing improved preservation of structural properties while improving PSNR of the recovered image. 

The performance boost on real dataset comes with a reduced performance on synthetic dataset. However when considering the broader perspective of deployment of these algorithms in real scenarios, such a performance trade-off between real and synthetic datasets is reasonable provided these methods retain their performance, when deployed in another domain. To evaluate this scenario, we refer to algorithms trained on aggregated dataset as being representative of strong baseline for further evaluation.  

\begin{table}[ht]
    \centering
    \caption{Algorithm Performance when trained on aggregated dataset} 
    \begin{adjustbox}{max width=\columnwidth}
    \begin{tabular}{l c c c c}
    \Xhline{3\arrayrulewidth} \hline \noalign{\vskip 1pt}
    Algorithm & SOTS-IN & SOTS-Out & NTIRE-19 & NTIRE-20 \\ 
    \Xhline{2\arrayrulewidth} \hline \noalign{\vskip 1pt}
    DuRN-US       & \textbf{26.80} / \textbf{0.95} & \textbf{29.59} / \textbf{0.91} & \underline{15.96} / \underline{0.61} & \underline{19.88} / \underline{0.69} \\
    FFA-Net       & 19.15 / 0.85 & 25.53 / 0.89 & 14.06 / 0.54 & 15.93 / 0.59 \\
    Wavelet       & 15.46 / 0.65 & 19.21 / 0.67 & 12.23 / 0.51 & 12.66 / 0.42 \\
    MSNet         & 24.38 / 0.90 & 26.79 / \underline{0.90} & 14.65 / 0.59 & 15.17 / 0.63 \\
    SNDN          & 22.94 / 0.88 & 25.49 / 0.89 & 14.66 / 0.59 & 18.65 / 0.67 \\
    GridDehazeNet & 22.87 / \underline{0.91} & \underline{27.18} / \textbf{0.91} & 13.97 / 0.56 & 17.02 / 0.68 \\
    PFFNet        & 23.39 / 0.87 & 25.80 / 0.87 & 13.50 / 0.48 & 14.77 / 0.57 \\ 
    Ours          & \underline{25.39} / 0.80 & 19.47 / 0.75 & \textbf{22.80} / \textbf{0.82} & \textbf{23.66} / \textbf{0.88} \\
    \Xhline{2\arrayrulewidth} \hline
    \end{tabular}
    \end{adjustbox}
    \begin{tablenotes}
    \scriptsize
    \item \textbf{Boldface} and \underline{Underlined} values represent best and second best results. 
    \end{tablenotes}
    \label{tab:tab_2}
\end{table}

\begin{figure*}[!ht]
    \centering
    \renewcommand{\tabcolsep}{1pt} 
    \renewcommand{\arraystretch}{1} 
    \begin{adjustbox}{max width=\textwidth}
    \begin{tabular}{cccccc}
    \includegraphics[width=0.166\textwidth, height=2.5cm]{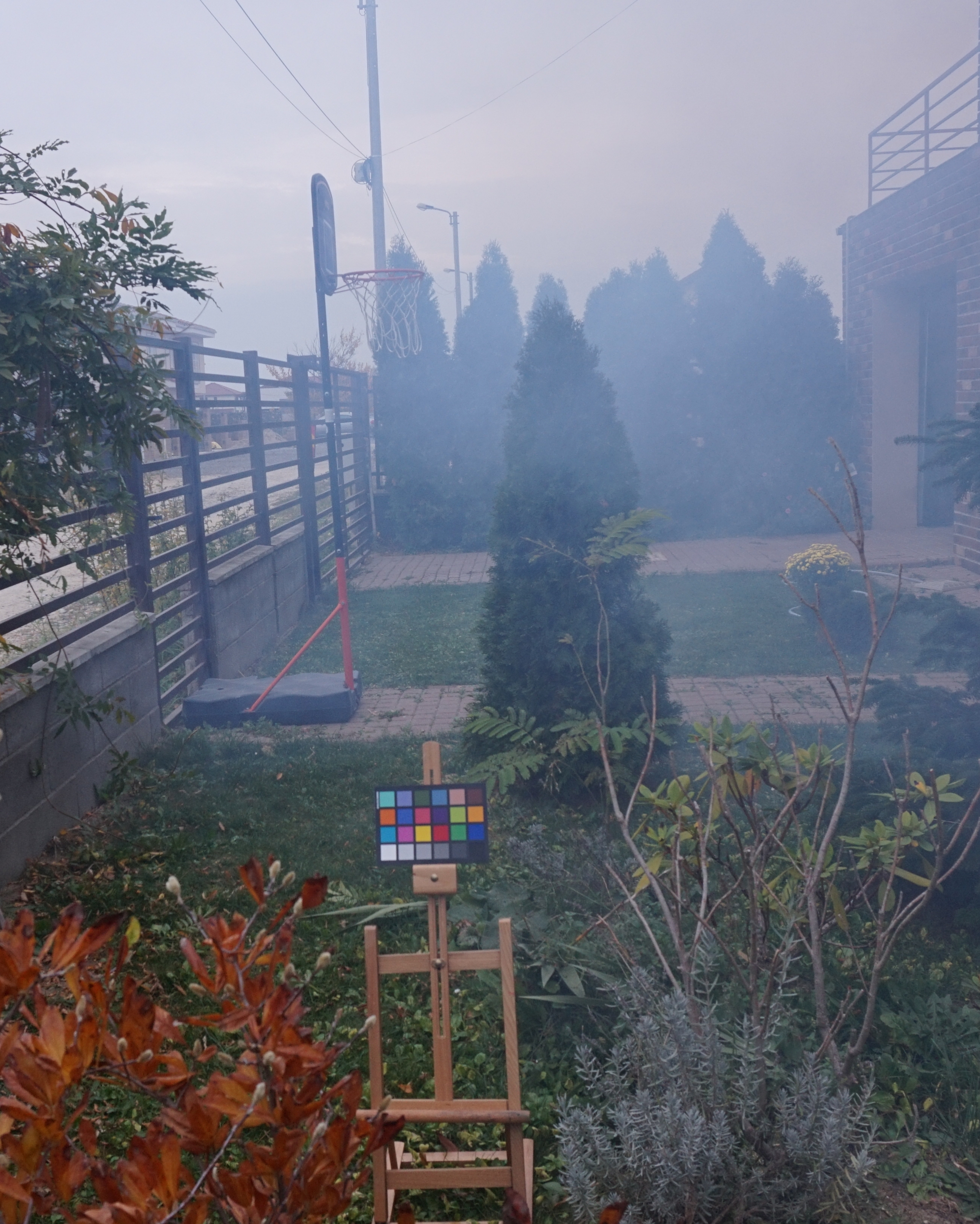} &
    \includegraphics[width=0.166\textwidth, height=2.5cm]{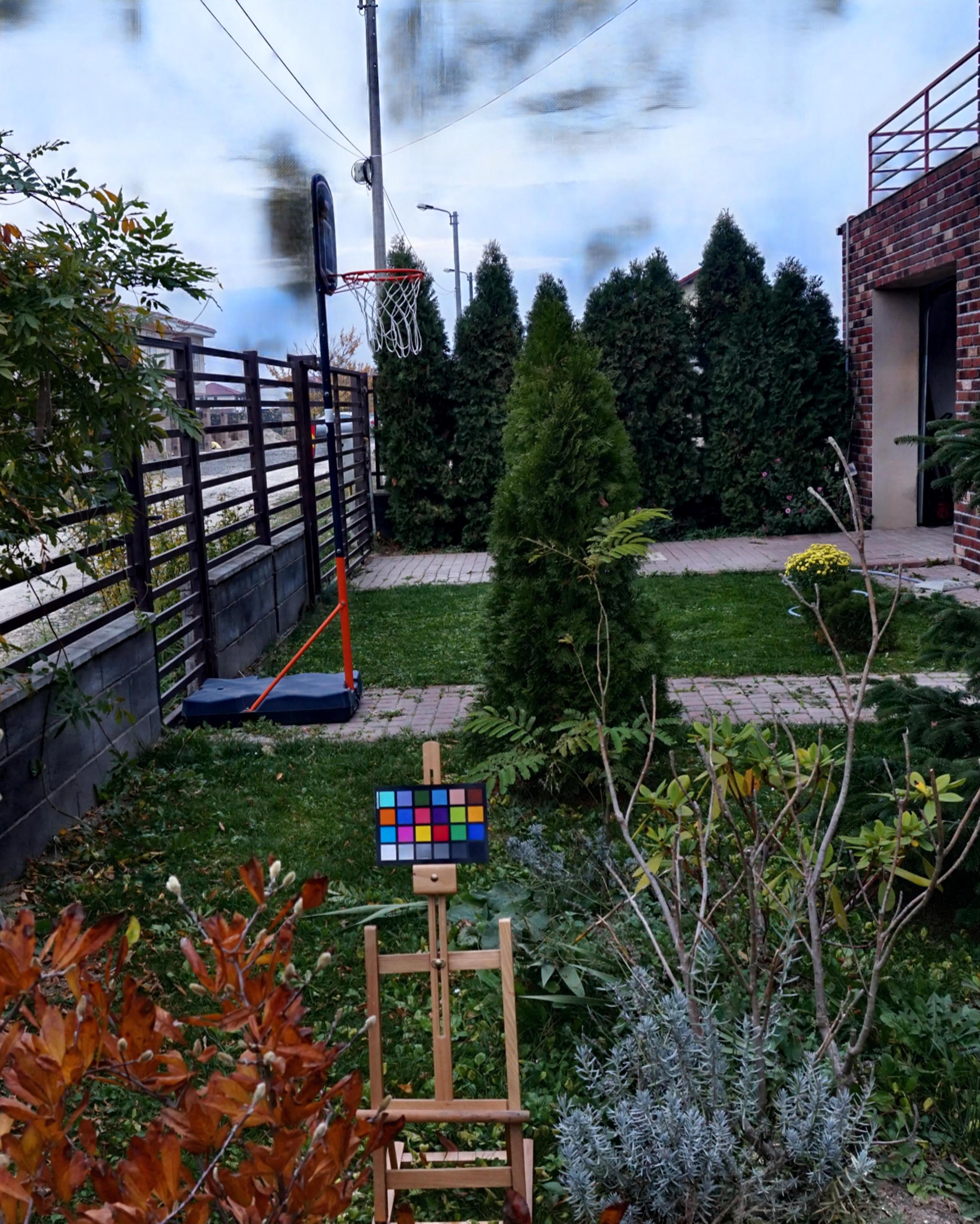} & 
    \includegraphics[width=0.166\textwidth, height=2.5cm]{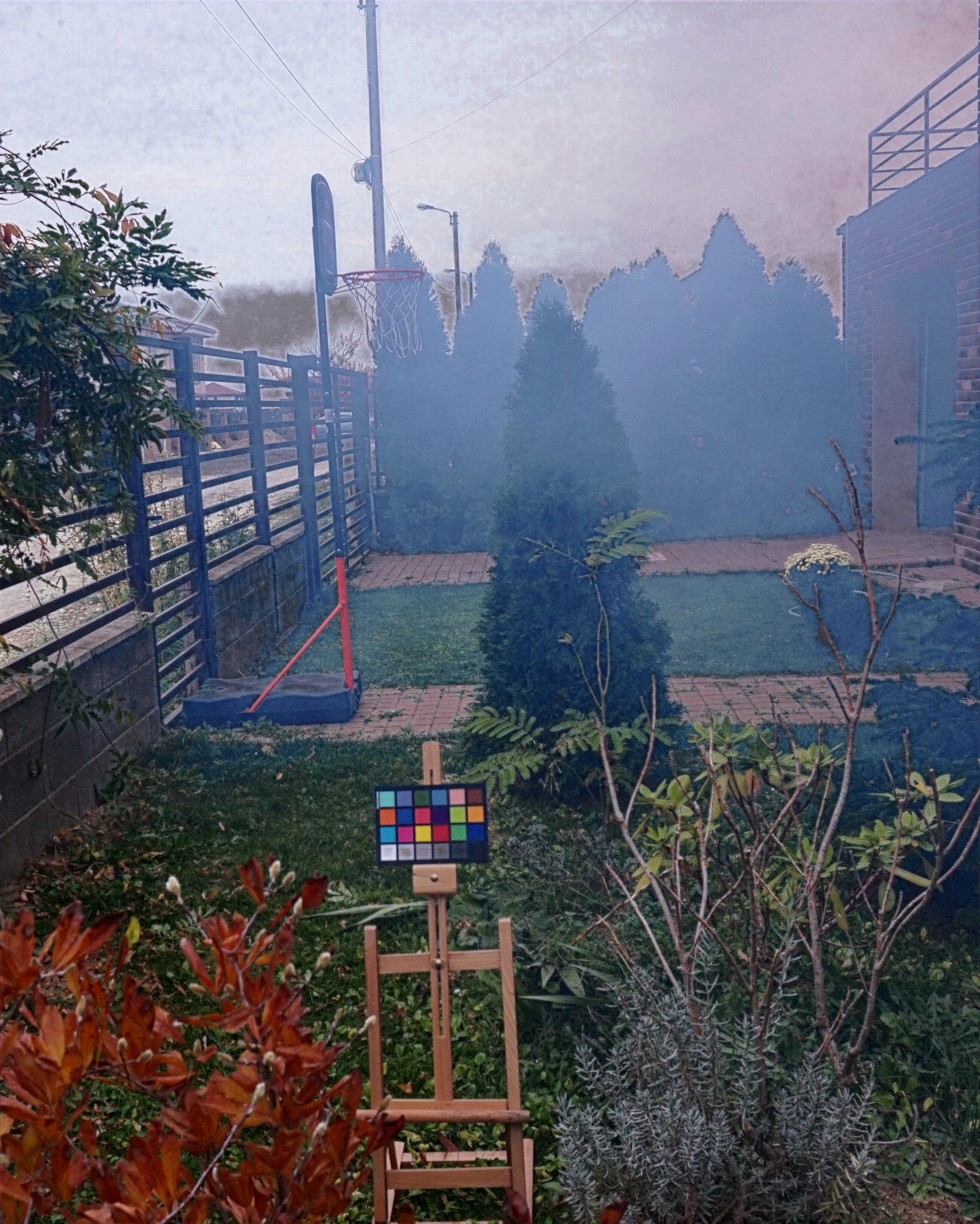} & 
    \includegraphics[width=0.166\textwidth, height=2.5cm]{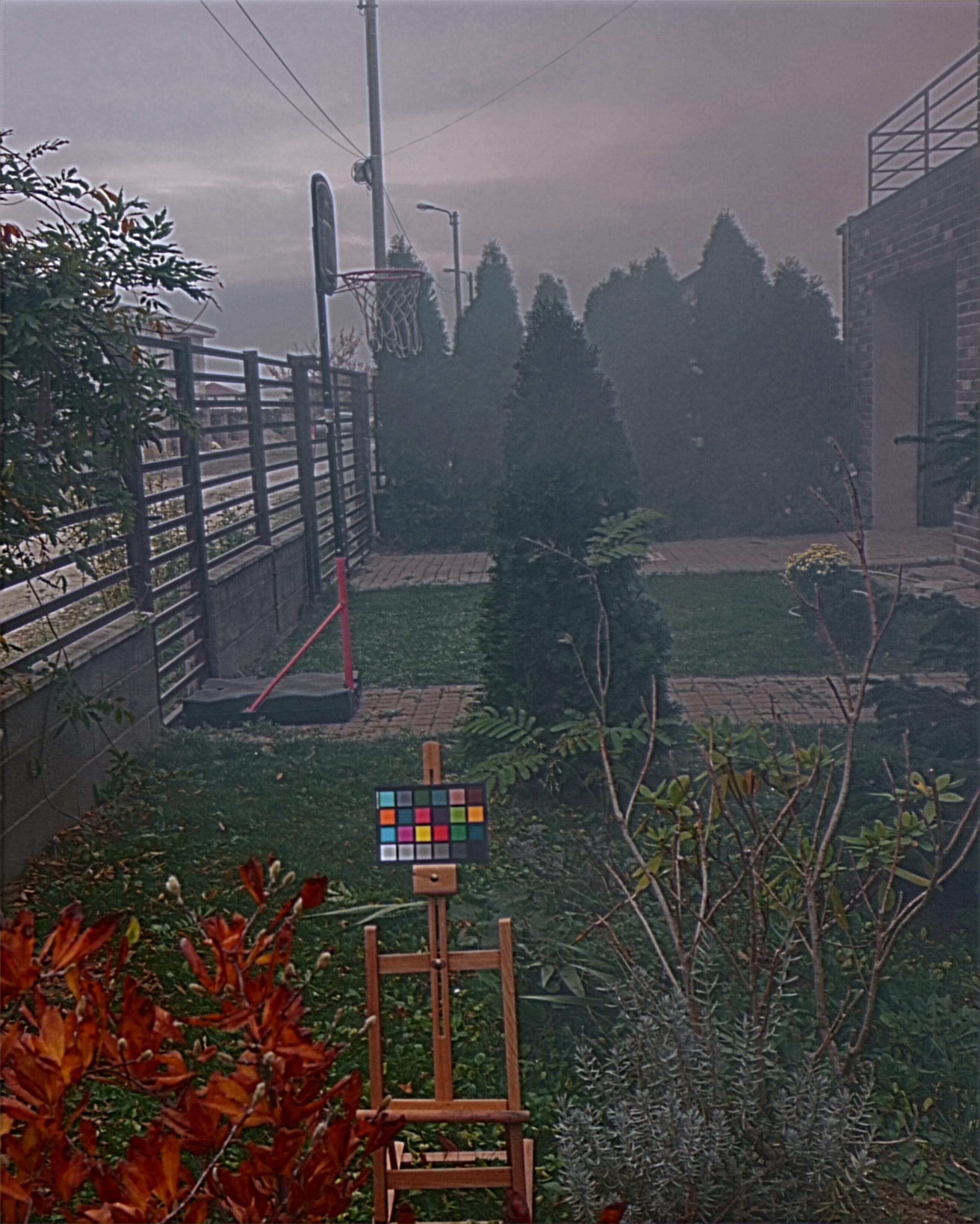} & 
    \includegraphics[width=0.166\textwidth, height=2.5cm]{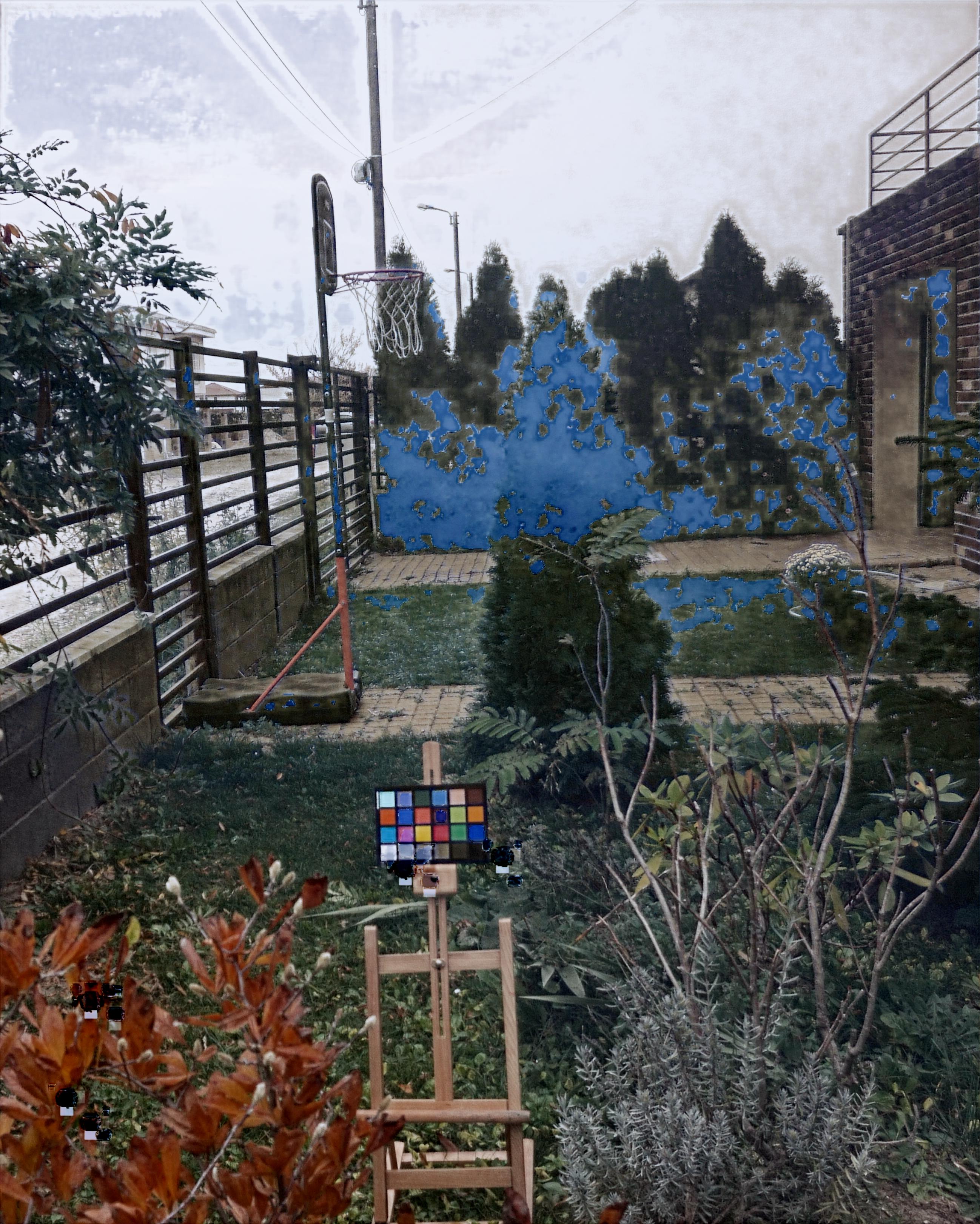} & 
    \includegraphics[width=0.166\textwidth, height=2.5cm]{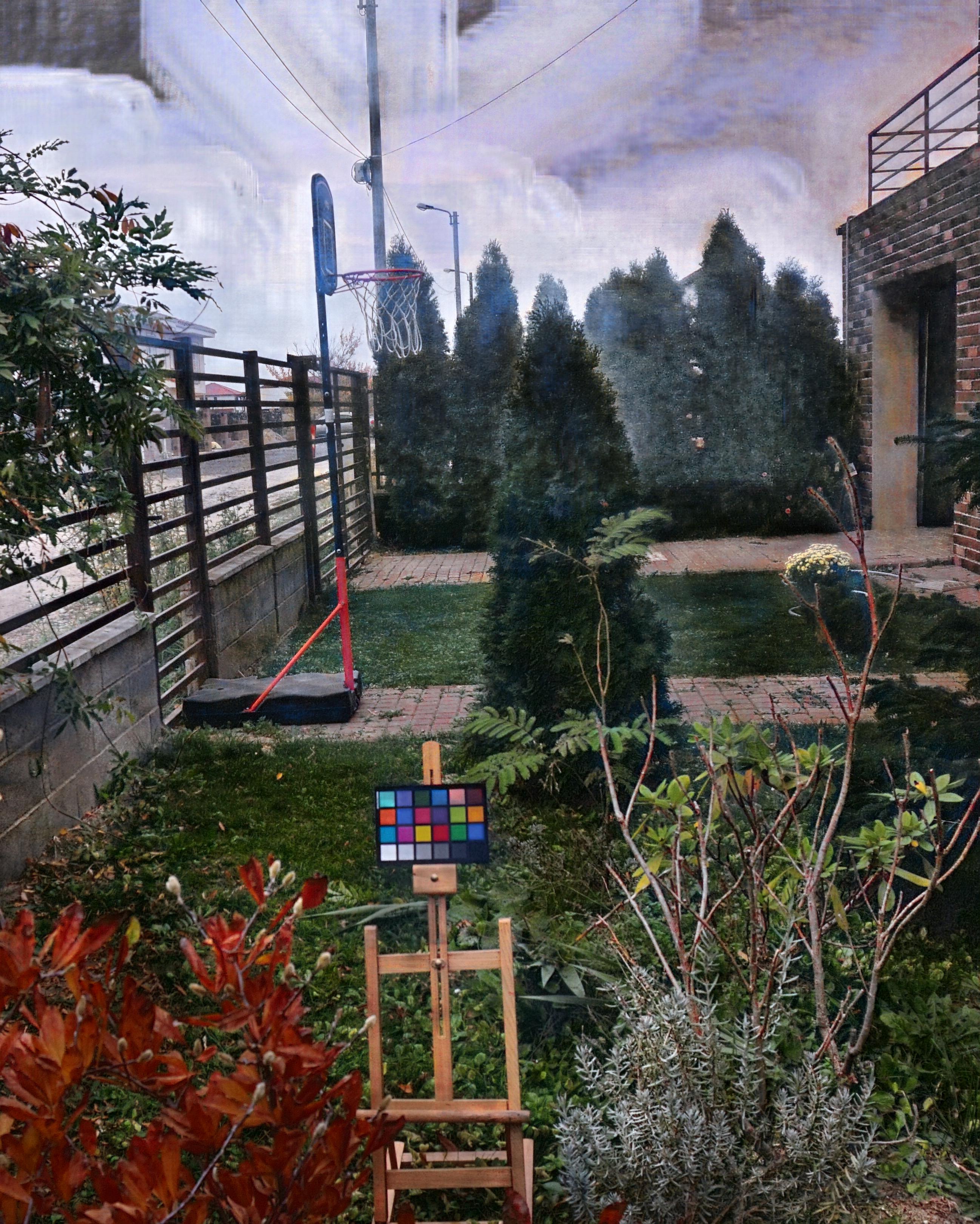} \\
    15.30 / 0.67 & 20.91 / 0.75  & 16.44 / 0.70  & 14.60 / 0.64  & 20.16 / 0.71  & 18.51 / 0.70\\
    Input Image  & DuRN-US       & FFA-Net       & Wavelet-UNet  & SNDN          & MSNet  \\  
    \includegraphics[width=0.166\textwidth, height=2.5cm]{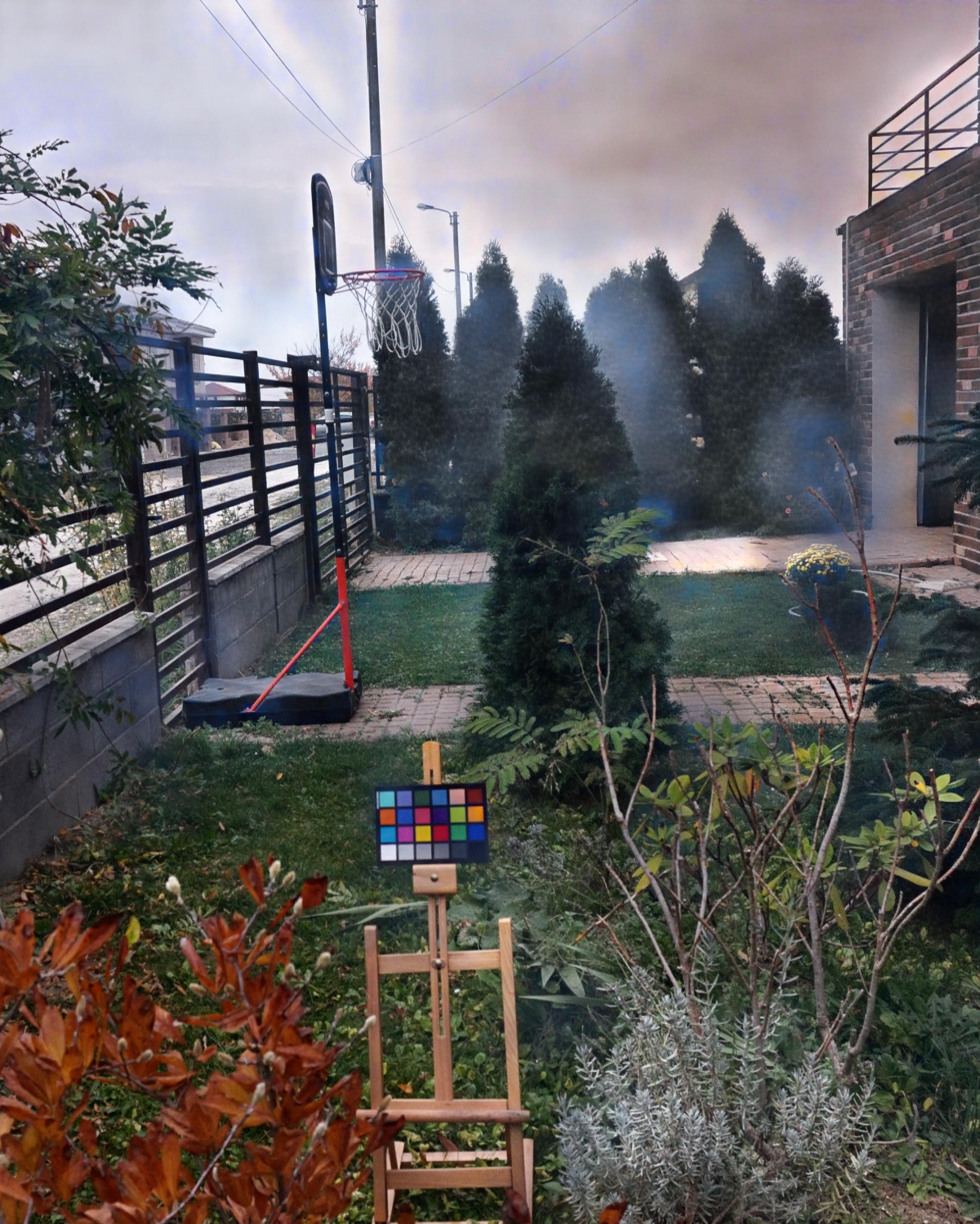} & 
    \includegraphics[width=0.166\textwidth, height=2.5cm]{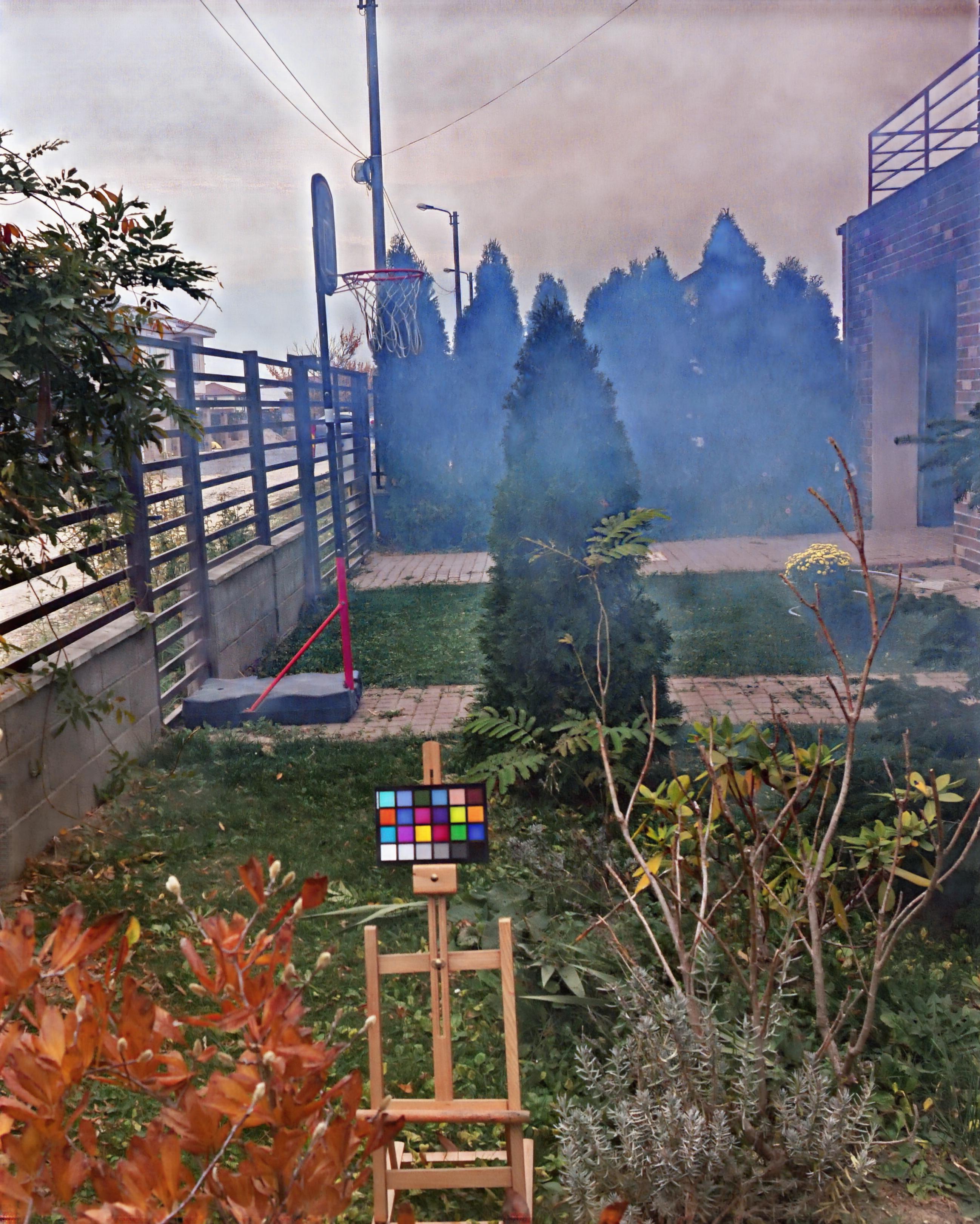} & 
    \includegraphics[width=0.166\textwidth, height=2.5cm]{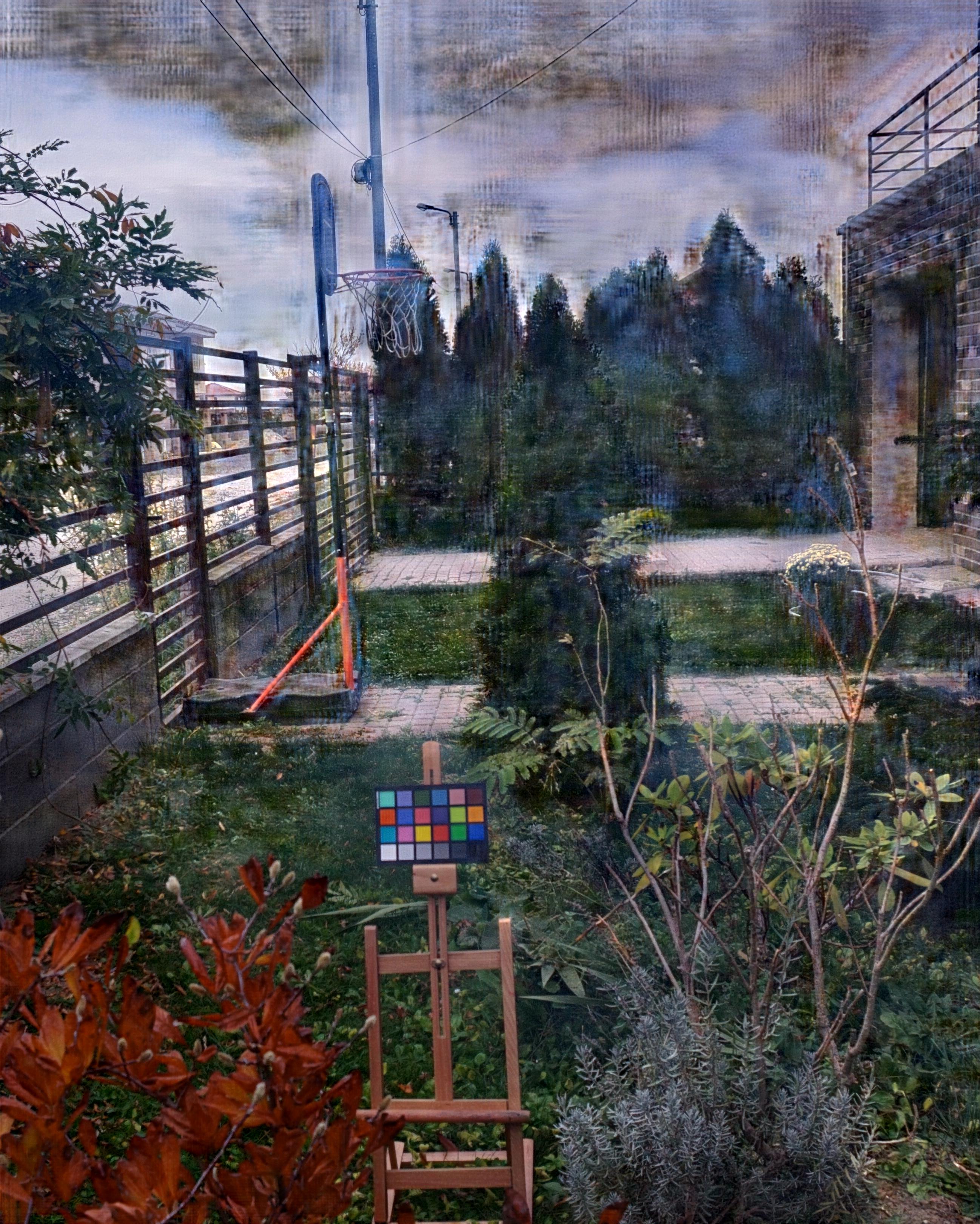} & 
    \includegraphics[width=0.166\textwidth, height=2.5cm]{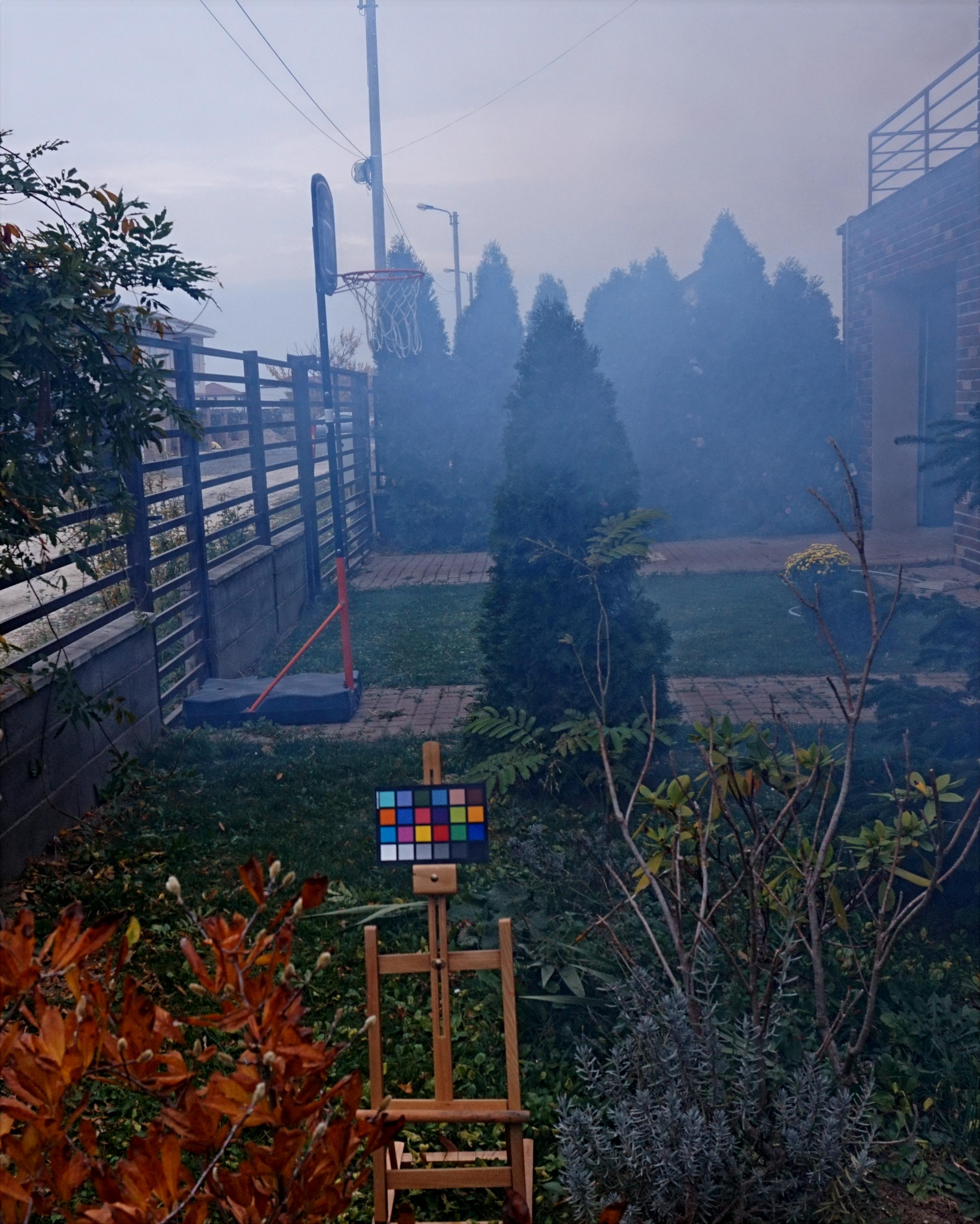} &
    \includegraphics[width=0.166\textwidth, height=2.5cm]{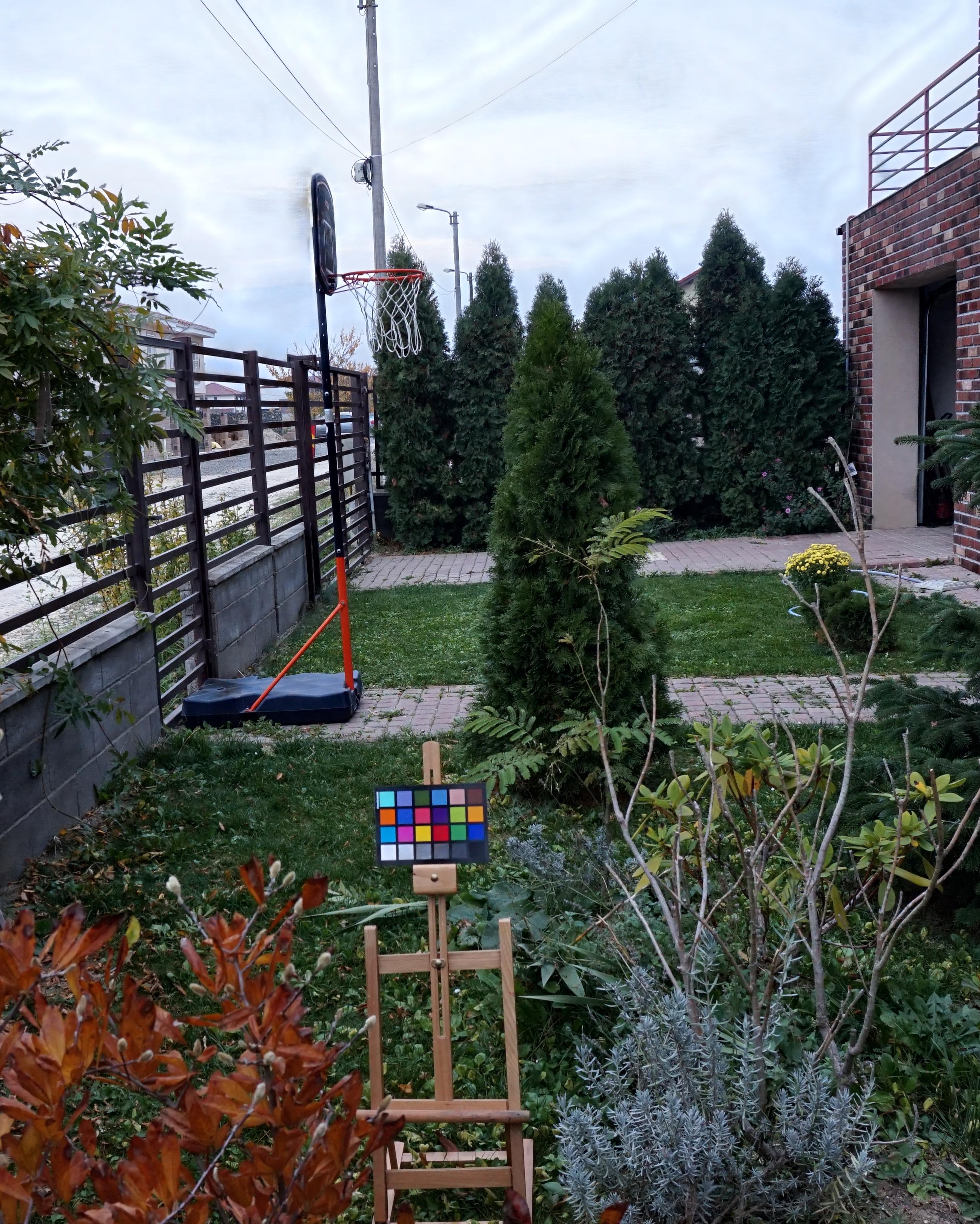} &
    \includegraphics[width=0.166\textwidth, height=2.5cm]{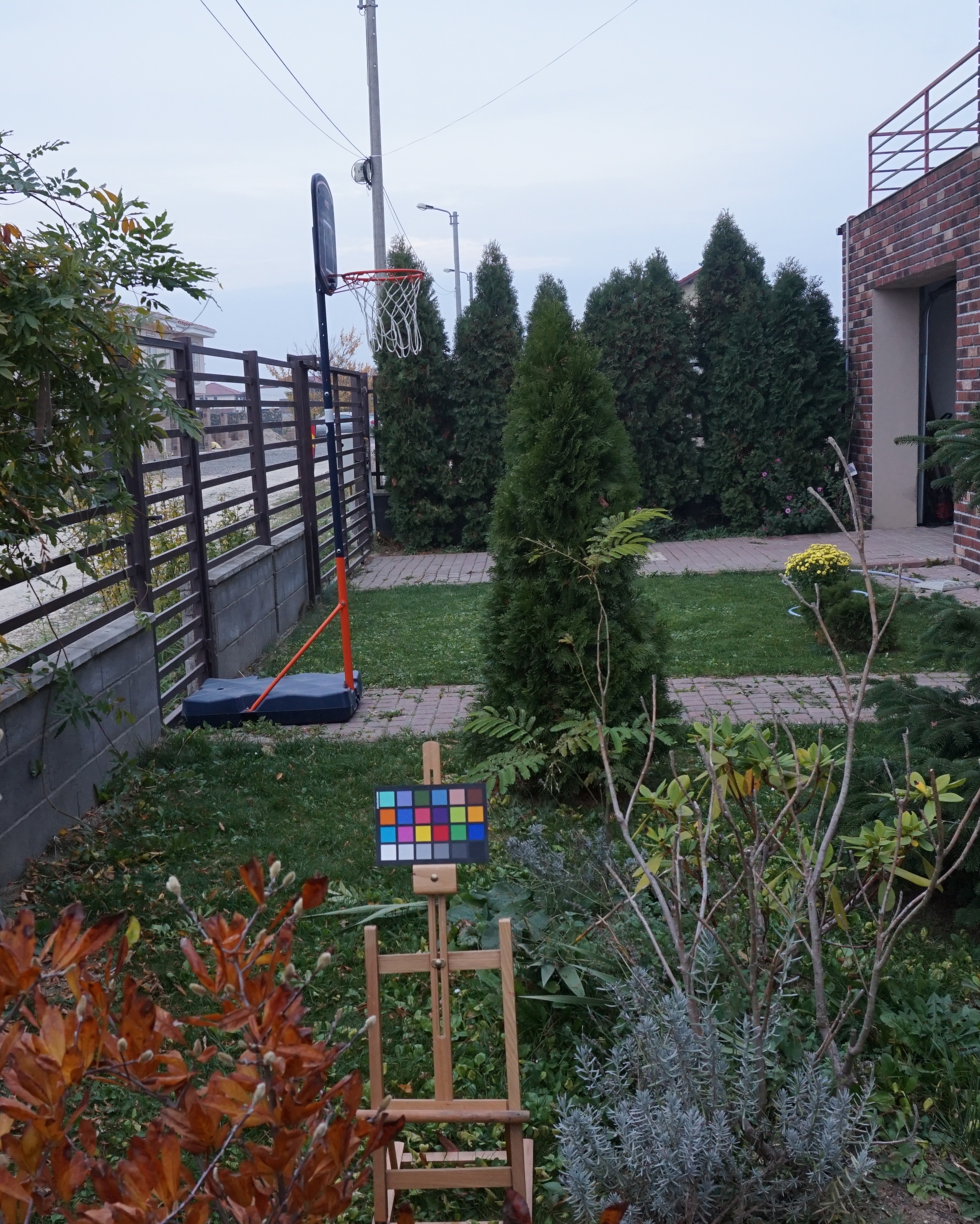} \\   
    19.15 / 0.74  & 16.20 / 0.71 & 16.13 / 0.65 & 16.13 / 0.65 & 23.26 / 0.82& \\
    GridDehazeNet & DA-Dehazing  & PFFNet       & Y-Net        & Ours        & Clean Image \\
    \end{tabular}
    \end{adjustbox}
    \caption{Blind Evaluation of different Dehazing algorithms on NTIRE-18 dataset}
    \label{fig:fig_5}
\end{figure*}

\textbf{Performance on datasets outside training distribution :}
To ascertain that higher performance of algorithms when trained on aggregated dataset ensures performance retention to unknown domains, we use Haze-RD and NTIRE-18 datasets for blind evaluation of strong baseline. We summarize numerical results in Tab. \ref{tab:tab_4} and visual results in Fig. \ref{fig:fig_5}. While the proposed framework retains its performance on NTIRE-18 dataset, performance of all algorithms on Haze-RD drops significantly. However performance drop in terms of PSNR is not substantial on NTIRE-18 for both proposed framework as well as strong baselines. Upon a visual examination of dehazed images, we observe that while performance in PSNR terms is mostly retained, prior works couldn't dehaze the image completely with some regions still affected by haze. Furthermore the structural properties of recovered objects is not retained. On contrary, the proposed framework was not only able to remove haze but also preserve color and structural properties of underlying objects to a substantial degree. Thereby demonstrating the effectiveness of proposed framework in unknown domains.


\begin{table}[ht]
    \centering
    \caption{Performance on datasets outside training distribution} 
    \begin{adjustbox}{max width=\columnwidth}
    \begin{tabular}{l c c}
    \Xhline{3\arrayrulewidth} \hline \noalign{\vskip 1pt}
    Algorithm & Haze-RD & NTIRE-18 \\ 
    \Xhline{2\arrayrulewidth} \hline \noalign{\vskip 1pt}
    DuRN-US       & 15.26 / \textbf{0.83} & \underline{18.85} / 0.71 \\
    FFA-Net       & 15.77 / \textbf{0.83} & 16.16 / 0.64 \\
    Wavelet       & \underline{16.30} / \underline{0.82} & 16.01 / 0.70 \\
    MSNet         & 14.42 / 0.80 & 17.42 / 0.66 \\
    SNDN          & 16.05 / \underline{0.82} & 18.08 / 0.70 \\
    GridDehazeNet & 14.58 / 0.81 & 18.26 / \underline{0.74} \\
    PFFNet        & 15.20 / 0.77 & 15.66 / 0.58 \\ 
    DA-Dehaze     & 16.21 / 0.78 & 16.28 / 0.67 \\
    Ours          & \textbf{21.42} / 0.81 & \textbf{24.14} / \textbf{0.80}\\
    \Xhline{2\arrayrulewidth} \hline
    \end{tabular}
    \end{adjustbox}
    \label{tab:tab_4}
\end{table}

\textbf{Ablation Studies : } We examine the effects of different strategies proposed in this paper for improving performance using NTIRE-19 and SOTS-IN datasets. The numerical results for different experiments are summarized in Tab. \ref{tab:tab_5}. We begin by evaluating a simple UNet architecture that acts as baseline model, upon which enhancements are performed. We observe using greedy local data augmentation technique (GLDA) significantly boosts performance of baseline model on both known and unknown datasets. We attribute this towards the ability of the network to focus explicitly on haze affected regions. To corroborate this observation, we progressively introduce SATA module with multi-scale feature aggregation (MSFA) and report continuous performance improvement both in terms of PSNR and SSIM, with SACA contributing more to SSIM whereas MSFA contributes towards improved PSNR. This validates the design choice to introduce these enhancements to preserve structural and feature properties respectively. While the PSNR and SSIM were considerably improved on NTIRE-19, the same wasnt observed for SOTS-IN dataset. Thus we examine the effect of using a HF prior based adversarial learning setup, which improves structural preservation across datasets but not the PSNR of recovered images. Subsequently we introduce an additional LF prior based discriminator and observe significant performance retention is achieved. This confirms our hypothesis of adding HF and LF prior based discriminators to preserve structural and color consistency within recovered images which wasnt possible when using a simple discriminator owing to weak supervision. 

\begin{table}[ht]
    \centering
    \caption{Effect of Different Strategies on Network Performance} 
    \begin{adjustbox}{max width=\columnwidth}
    \begin{tabular}{l c c}
    \Xhline{3\arrayrulewidth} \hline \noalign{\vskip 1pt}
    Algorithm & SOTS-IN & NTIRE-19  \\ 
    \Xhline{2\arrayrulewidth} \hline \noalign{\vskip 1pt}
    Baseline            & 18.08 / 0.51 & 11.05 / 0.34 \\
    GLDA                 & 21.53 / 0.74 & 15.32 / 0.53 \\
    SACA                & 22.07 / 0.82 & 16.70 / 0.59 \\
    MSFA                & 26.75 / 0.85 & 17.18 / 0.62 \\
    Simple Discriminator& 28.52 / 0.89 & 17.72 / 0.69 \\
    HF prior            & 32.87 / 0.94 & 19.02 / 0.72 \\
    LF and HF prior     & 38.91 / 0.98 & 19.47 / 0.75 \\
    \Xhline{2\arrayrulewidth} \hline
    \end{tabular}
    \end{adjustbox}
    \label{tab:tab_5}
\end{table}

\section{Conclusion}
In this paper, we focused on the dual challenge of domain and haze distributions that significantly reduces the performance of dehazing models. To overcome this, we first proposed a spatially aware channel attention mechanism integrated within a CNN for increasing the receptive field and utilized local data augmentation to simulate non-uniform haze regions. We then trained the proposed network within an adversarial framework that uses high and low frequency components as priors to determine whether a given image is real or fake. This is shown to improve performance retention in unknown domains. We perform extensive experiments to demonstrate the effectiveness of different methods aiding the proposed method to achieve SoTA and domain invariant performance,

\section{ Acknowledgments}
This research was supported in part by KAIST-KU Joint Research Center, KAIST, Korea (N11200035) and by Next-Generation Information Computing Development Program through the National Research Foundation of Korea (NRF) funded by the Ministry of Science, ICT (NRF-2017M3C4A7069369). We gratefully acknowledge the GPU donation from NVidia used in this research.

\bibliography{references.bib}
\end{document}